\documentclass[journal]{IEEEtran}

\usepackage{amsmath,subfig,color,url}
\usepackage{cite}
\usepackage[T1]{fontenc}
\usepackage[utf8]{inputenc}
\usepackage{tabularx}
\usepackage{colortbl}
\usepackage{hhline}
\usepackage{amsmath}
\usepackage{bm}
\usepackage[table]{xcolor}
\usepackage{amssymb}
\usepackage[inline]{enumitem}
\usepackage{mathtools}
\usepackage{booktabs}
\usepackage[ruled]{algorithm2e}
\usepackage{float}
\usepackage{cite}
  \usepackage[pdftex]{graphicx}
  \usepackage{epstopdf}
  \graphicspath{{./figure/}}
\ifCLASSINFOpdf
\else
\fi

\usepackage{stfloats}
\usepackage{multirow}
\begin{document}
\title{Phrase-based Image Captioning with \\ Hierarchical LSTM Model}

\author{Ying~Hua~Tan
        and~Chee~Seng~Chan% <-this % stops a space
\thanks{Y.H. Tan and C.S. Chan are with the Center of Image and Signal Processing, Department of Artificial Intelligence, Faculty of Computer Science and Information Technology, University of Malaya, Kuala Lumpur,
50603 MALAYSIA. e-mail: \{tanyinghua@siswa.um.edu.my; cs.chan@um.edu.my\}}% <-this % stops a space
}


\markboth{ACCV2016 Extension}%
{Shell \MakeLowercase{\textit{et al.}}: Bare Demo of IEEEtran.cls for IEEE Journals}



\maketitle

\begin{abstract}
Automatic generation of caption to describe the content of an image has been gaining a lot of research interests recently, where most of the existing works treat the image caption as pure sequential data. Natural language, however possess a temporal hierarchy structure, with complex dependencies between each subsequence. In this paper, we propose a phrase-based hierarchical Long Short-Term Memory (phi-LSTM) model to generate image description. In contrast to the conventional solutions that generate caption in a pure sequential manner, our proposed model decodes image caption from phrase to sentence. It consists of a phrase decoder at the bottom hierarchy to decode noun phrases of variable length, and an abbreviated sentence decoder at the upper hierarchy to decode an abbreviated form of the image description. A complete image caption is formed by combining the generated phrases with sentence during the inference stage. Empirically, our proposed model shows a better or competitive result on the Flickr8k, Flickr30k and MS-COCO datasets in comparison to the state-of-the art models. We also show that our proposed model is able to generate more novel captions (not seen in the training data) which are richer in word contents in all these three datasets.
\end{abstract}


% \fi
\IEEEpeerreviewmaketitle

\section{Introduction}
Automatic caption or description generation from images is a challenging problem that requires a combination of visual information and linguistic. In other words, it requires not only complete image understanding, but also sophisticated natural language generation \cite{bernardi2016automatic}. This is what makes it such an interesting task that has been embraced by both the computer vision and natural language processing communities. 

Over the past few years, one of the most common frameworks applied in this line of research is a neural network model composed of two sub-networks \cite{mao2014deep,Vinyals2015,karpathy2015deep,kiros2014unifying,donahue2015long}, where a convolutional neural network (CNN) is used to encode the image into a feature representation; while a recurrent neural network (RNN) is applied to decode it into a natural language description. In particular, the Long Short-Term Memory (LSTM) model \cite{hochreiter1997long} has emerged as the most popular RNN architecture, as it has the ability to capture long-term dependency and preserve sequence. Recently, many variants of this framework were introduced and achieved good results, such as those with attention mechanism \cite{xu2015show, yang2016review, fu2016aligning} and attributes \cite{wu2017image, you2016image}. However, we notice that most of these works decode image caption in a fully sequential word-by-word basis. Although sequential model is appropriate for processing sentential data, it does not capture any other syntactic structure of language at all. 

In fact, natural language is one of those sequential data that has temporal hierarchy, with information spread out over multiple time-scales \cite{hermans2013training}. Consider English as an example, the lowest level with the shortest time-scale is characters, followed by words, phrases, clauses, sentences to documents. Therefore, it is undeniable that sentence structure is one of the prominent characteristics of language, and  Victor Yngve, an influential contributor in linguistic theory stated in 1960 that ``\textit{language structure involving, in some form or other, a phrase-structure hierarchy, or immediate constituent organization}''\cite{yngve1960model}. Hence, forcing a generative model trained on flat sequences to generate a high-level structure locally in a step-by-step basis often results in limited performance \cite{serban2017hierarchical}. For image caption in particular, there are at least two levels of structure observed from the human annotated captions in the public datasets we experimented. Within each of the caption, there are several phrases that describe the objects in an image. These phrases have equal time-scale at the word level, and they are conditioned on both the image and short-term language structure during decoding. Thus, previous sequence of caption excludes the phrase itself, encoded in the long term memory is redundant in its generation process. On the other hand, the structure of caption across these phrases is more inter-dependent, and requires both the image and all previous sequences as context to generate a correct description. 

\begin{figure}[t]
	\includegraphics[height=0.4\linewidth, width=0.9\linewidth]{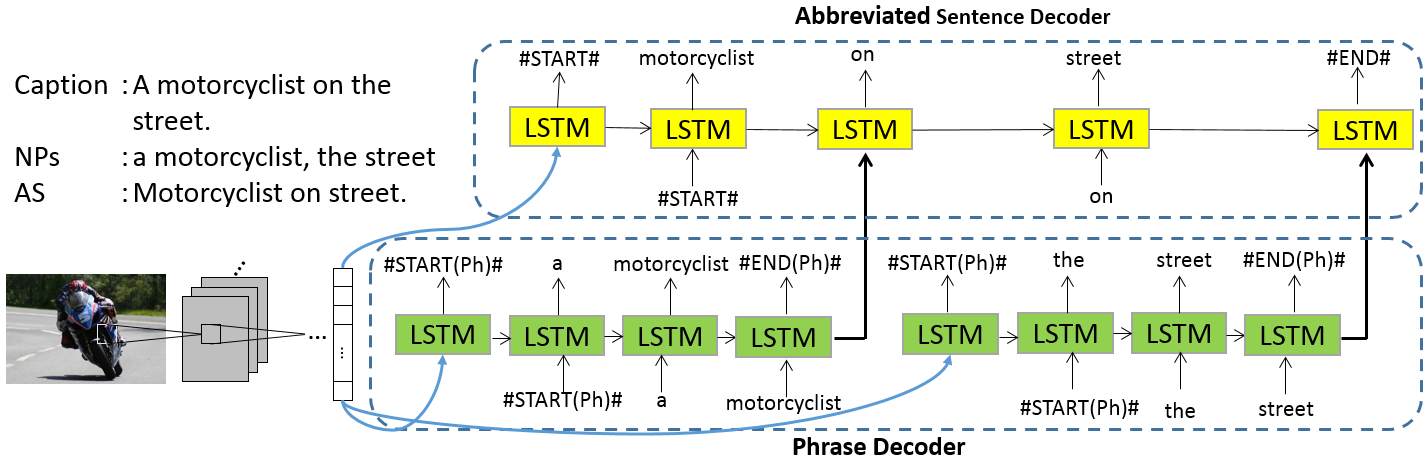}
	\centering
	\caption{The overall architecture of the phi-LSTM model. It consists of a phrase decoder at the lower hierarchy and an AS decoder at the upper hierarchy.} \vspace{-.1in}
	\label{fig:overall}
\end{figure}

In this paper, we would like to investigate the capability of a phrase-based image captioning model that incorporated the observed structure in its modeling, as compared to a similar model trained on flat sequences. To this end, we design a phrase-based hierarchical LSTM model, namely {\bf phi-LSTM} that consists of a phrase decoder and an abbreviated sentence (AS) decoder to generate image description from phrase to sentence. As illustrated in \figurename~\ref{fig:overall}, given an image encoded with the CNN, the phrase decoder is employed to decode noun phrases (NPs) (i.e a motorcyclist, the street) that describe the dominant entities within the image, using words as atomic unit. At the same time, the phrase decoder also encodes each of the NP into a compositional vector representation, which serves as an input to the AS decoder at the upper hierarchy. Hence, the NPs will have equal time step as the remaining words at the sentence level (i.e on). Then, the AS decoder will decode an abbreviated form of the caption, which is made up of the last word of each NP (i.e motorcyclist, street) and the remaining words that connect the phrases (i.e on). A complete image caption (i.e A motorcyclist on the street) is formed by combining the generated phrases with sentence gradually during beam search at the inference stage. Empirically, our proposed model shows a better or competitive results on Flickr8k \cite{rashtchian2010collecting}, Flickr30k \cite{young2014image} and MS-COCO \cite{lin2014microsoft} datasets in comparison to the state-of-the art models.

As a summary, our contributions are two-folds:
\begin{enumerate}
	\item We propose a novel phrase-based hierarchical LSTM model to decode image caption from phrase to sentence.
	\item We show that the image caption generated with phi-LSTM is more accurate, novel (not seen in training data), and richer in word content. %This is because the local statistic of language that dominates the subsequences (NPs) are removed from the global sequence (AS), allowing more variability to be injected when decoding the AS. 
\end{enumerate}

A preliminary version of this work was presented in \cite{tan2017phi}, whereas the present work adds to the initial version in significant ways. First, the phrase selection objective is replaced with prediction of the last word of each NP with the AS decoder for training simplicity. Secondly, we introduce length normalization during the inference stage at both phrase and sentence level, in order to generate longer caption. Thirdly, we further improve the outputs of parsing tool with a phrase refinement strategy. Finally, considerable new analysis and intuitive explanations are added to our results. We also extend our experiment to include the MS-COCO dataset \cite{lin2014microsoft}, and evaluate our results on four additional evaluation metrics (i.e METEOR \cite{banerjee2005meteor}, ROUGE \cite{lin2004rouge}, CIDEr \cite{vedantam2015cider} and SPICE \cite{anderson2016spice}).

\section{Related Works}

The image description generation approaches are differed in terms of
\begin{enumerate*}[label={\roman*)}]
	\item how the context in which the description is derived from is represented, and
	\item how a sentence is generated.
\end{enumerate*}

\subsection{Context Representation}
To encode visual information, earlier works rely on multiple visual detectors and classifiers to capture different aspects of an image, such as objects, attributes, relations and scene \cite{farhadi2010every,kulkarni2011baby,li2011composing,yang2011corpus,mitchell2012midge,kuznetsova2012collective,kuznetsova2014treetalk}. The outputs of these detectors and classifiers usually form a set of tuples \cite{farhadi2010every,kulkarni2011baby,li2011composing,yang2011corpus,mitchell2012midge}, in which the description is built upon. Such approach generally fixes the number of classes for each aspect of the image. Since the unprecedented success of CNN in image classification and object detection tasks, a growing number of works start to use different variants of CNN to encode a whole image \cite{socher2014grounded, mao2014deep, donahue2015long, devlin2015language, kiros2014multimodal, kiros2014unifying, Vinyals2015, lebret2015phrase, jia2015guiding, xu2015show, yang2016review, you2016image, ushiku2015common}, or multiple image regions \cite{karpathy2014deep, karpathy2015deep, fang2015captions,fu2016aligning, wu2017image}. Given the CNN encoded image and its description, many works train a multimodal embedding space using various language model \cite{socher2014grounded,mao2014deep,donahue2015long,jia2015guiding, karpathy2014deep,karpathy2015deep, kiros2014multimodal, kiros2014unifying, lebret2015phrase,ushiku2015common, Vinyals2015,xu2015show, yang2016review, you2016image} to decode image caption. Alternatively, Fang et al.\cite{fang2015captions}, Wu et al. \cite{wu2017image} and You et al.\cite{you2016image} train a set of ``visual word detectors'' on the training data to encode image into a semantic space.

Besides that, there are works that rely on retrieval approach to generate image description. By retrieving and re-ranking the caption of similar images from the training sets \cite{ordonez2011im2text, hodosh2013framing, socher2014grounded, karpathy2014deep, devlin2015language}, a query image can be described with human written caption that is most relevant to its content. However, this method is incapable of describing an image with unseen composition of objects correctly. Thus, some of the works in this line of approach retrieve a set of tuples \cite{farhadi2010every} or text snippets \cite{kuznetsova2012collective, gupta2012choosing, kuznetsova2014treetalk} to form and re-rank novel captions.

\subsection{Description Generation}
Given various contexts described above, several approaches are developed to generate image description, which are
\begin{enumerate*}[label={\roman*)}]
	\item template-based,
	\item composition-based, and
	\item language model-based.
\end{enumerate*}

\subsubsection{Template-based}
This approach generates sentence using a pre-defined template with open-slots to be filled with image entities \cite{farhadi2010every,kulkarni2011baby,yang2011corpus,gupta2012choosing}. It is mostly used by works that represent visual content as a set of tuples. Description generated this way is usually syntactically correct, but rigid and not flexible.

\subsubsection{Composition Method}
This approach stitches up text snippets retrieved \cite{kuznetsova2012collective,kuznetsova2014treetalk} or entities detected \cite{li2011composing,mitchell2012midge} to form an image description. It requires sophisticated pre-defined rules to decide the set of text snippets or entities to be used for generating a complete caption, their orders and the gluing words in between them. Description generated in such manner is broader and more expressive compared to the template-based approach, but is also computationally expensive at test time due to its nonparametric nature.

\subsubsection{Language Model-based}
Most recent works jointly embed image and language into a multimodal embedding space with neural network based language model to generate image caption \cite{kiros2014multimodal, kiros2014unifying, mao2014deep, karpathy2015deep, donahue2015long, vinyals2017show}. For instance, Kiros et al. \cite{kiros2014multimodal} proposed a multimodal log-bilinear neural language model which is biased by image feature to decode image caption. Mao et el. \cite{mao2014deep} and Karpathy \& Li \cite{karpathy2015deep} used RNN to decode caption of varying length, while LSTM was implemented in \cite{donahue2015long, Vinyals2015, jia2015guiding, wu2017image} to decode image description from their respective context. For example, Jia et al. \cite{jia2015guiding} used both CNN encoded image and semantic embedding learned with normalized Canonical Correlation Analysis as inputs to their LSTM decoder. Moreover, Xu et al. \cite{xu2015show}, Fu et al. \cite{fu2016aligning} and Yang et al. \cite{yang2016review} incorporated attention mechanism with the LSTM decoder to attend to various parts of image during the caption generation process. On the other hand, You et al. \cite{you2016image} implemented attention mechanism over semantic space instead of multimodal space when generating image caption.
%whereas the former joint image feature with hidden state of RNN before prediction of next word in sequence, while the later bias the initial input of RNN with image feature. Yang et al. \cite{yang2016review} also proposed a variant of attention mechanism, where \textit{T} time step of LSTM is used to review CNN features attended to various regions, and the outputs of all review LSTMs are used to generate image caption with LSTM decoder. 

\subsection{Relation to Our Work}

Similarly, our model employ the LSTM to decode image caption using CNN encoded image as context. However, instead of using tokenized words as atomic unit to a pure sequential LSTM, we introduce a hierarchical LSTM structure to decode image description from phrase to sentence. Thus, the input of our model at sentence level is a sequence of combination of words and phrases. 

Our work is different from the phrase-based approaches that use retrieval of text snippets paired with template or composition method to generate caption \cite{kuznetsova2012collective, gupta2012choosing, kuznetsova2014treetalk}, as we do not rely on retrieval. Other phrase-based approaches place more emphasize on phrase learning and use a simple language model to decode sentence. For example, Lebret et al. \cite{lebret2015phrase} and Ushiku et al. \cite{ushiku2015common} extracted various types of phrase from image description. The former trained phrase relevancy with image with negative sampling, and decoded a sequence of phrases using a tri-gram language model conditioned on the chunking tag of each phrase. The latter proposed a subspace-embedding method for phrase learning and generated sentence from estimated phrases using a combinatorial optimization. Our work differs from them in terms of 
\begin{enumerate*}[label={\roman*)}]
	\item the type of phrase extracted, 
	\item phrase learning approach, and 
	\item sentence decoding method. 
\end{enumerate*}
First, we only extract NPs with intuition of having each phrase equivalent to an entity within the image. Moreover, we train both of our phrase and AS decoder using the LSTM, which are linked hierarchically as shown in \figurename~\ref{fig:overall}, such that our phrase representation is learned from the backpropagation of AS decoder at sentence level. Lastly, we generate a complete caption by decoding AS while progressively replace the inferred noun with generated phrases. % based on conditions stated in Section \ref{sec:generation}. 

A very recent work published, Skeleton-Key \cite{wang2017skeleton} is currently the closest work to ours. They designed a course-to-fine image caption decoder consists of two submodels, where Skel-LSTM learns to generate skeleton sentence made up of original caption with each NP replaced with its last word, while Attr-LSTM learns to decode the NPs. Although their model seems to resemble ours, there is still a distinct difference on how we links these two submodels. First, Wang et al. \cite{wang2017skeleton} designed a top-down model, where skeleton sentence is first generated, followed by decoding each skeletal word to form the attribute sub-sequences. On the contrary, our model is a bottom-up approach where NPs are first generated before a complete description. Secondly, during the testing stage, they used a length factor to control the length of generated caption manually, whereas our phrase indication objective and the normalized log probability of each NP candidate govern the length of the generated caption automatically. Finally, we do not implement attention mechanism, as this is beyond the scope of this paper.
%the scope of this paper is to analyze the characteristic of image caption generated by modeling the hierarchical nature language data, as compared to caption generated with pure sequential based language model

\section{Phi-LSTM Architecture}
Given an image-sentence pair, NPs that are equivalent to the entities within the image are first chunked from the sentence (S), using a phrase chunking algorithm described in Section \ref{sec:parsing}. Then, an AS is formed by replacing each NP in the caption with the last word of the chunked phrase as shown in the example below:

\begin{description}[leftmargin=3em,style=nextline, font=\small]
\small
\item	[S:] \textit{\underline{The man} in \underline{the gray shirt} and sandals is pulling \underline{the large tricycle}.}
\item	[NPs:] \textit{the \underline{man}, the gray \underline{shirt}, the large \underline{tricycle}}
\item	[AS:] \textit{Man in shirt and sandals is pulling tricycle.}
\end{description}

We decompose each caption in the training data into an AS-NPs pair, such that the AS and NPs are processed with two decoders that are linked hierarchically. The decomposition alters the order of sequence in the human annotated caption, and thus we have different ground truth sequence (GTS) during the training stage as compared with the conventional RNN models. The GTS of our phrase decoder is the NPs, while the GTS of our AS decoder is the AS, as described below.

\subsection{Phrase Decoder}
The phrase decoder in this work has two roles, which are 
\begin{enumerate*}[label={\roman*)}]
\item decodes an image representation into multiple NPs that describe the entities within the image, and 
\item encodes each of the NP into a compositional vector representation, which serves as an input to the AS decoder. 
\end{enumerate*}

\begin{figure}[t]
	\includegraphics[height=0.4\linewidth, width=0.85\linewidth]{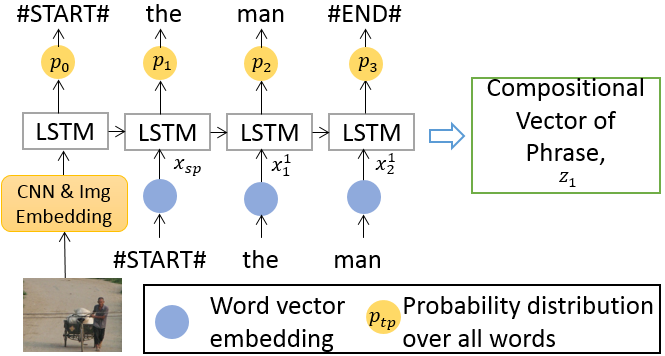}
	\centering
	\caption{The phrase decoder is trained to generate NPs and encode each NP into a compositional vector.}\vspace{-.1in}
	\label{fig:phrase_rep}
\end{figure}

\begin{figure*}[!t]
	\centering
	\includegraphics[height=0.25\linewidth, width=0.7\linewidth]{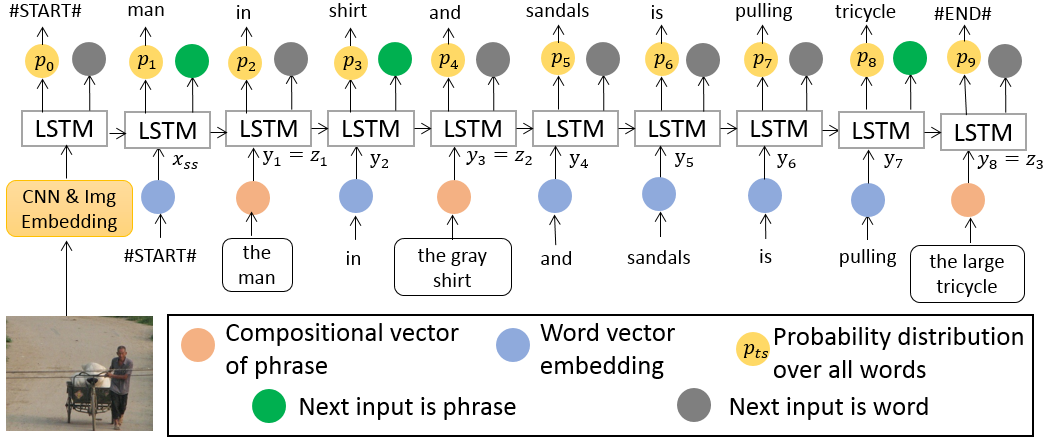}
	\caption{Abbreviated sentence decoder: The input sequence is a complete caption, with each NP occupies only one time step, while the ground truth sequence is the AS of the caption. It also predicts whether the next input is a phrase or a word.}\vspace{-.1in}
	\label{fig:sent_RNN}
\end{figure*}

Given an image \textit{I}, a CNN pre-trained on ImageNet \cite{deng2009imagenet} classification task is applied to encode an image into a \textit{D}-dimensional image feature, which is then transformed into a \textit{K}-dimensional vector with image embedding matrix, $\mathbf{W_{ip}} \in \mathbb{R}^{K \times D}$ and bias $\mathbf{b_{ip}} \in \mathbb{R}^K$. Then, a LSTM model similar to \cite{Vinyals2015} is used to decode it into each of the NPs.

To train the LSTM model to decode \textit{i}-th NP of length $L_i$, the embedded image feature, followed by a start-word token $\mathbf{x_{sp}} \in \mathbb{R}^{K}$ indicates the translation process, and each word in the NP are input to a LSTM block in a step-by-step manner, as shown in \figurename~\ref{fig:phrase_rep}. Hence, the phrase decoder inputs $\mathbf{x_{t_p}^i}$ at each time step of phrase, $t_p$ are:
\begin{equation}
\mathbf{x_{t_p}^i} =
\begin{cases}
\mathbf{W_{ip}} \text{CNN}(I) + \mathbf{b_{ip}}~, & \text{for}\ t_p=-1 \\
\mathbf{x_{sp}}~, & \text{for}\ t_p=0 \\
\mathbf{W_{ep}}w_{tp}^i~, & \text{for}\ t_p = {1...L_i}~,
\end{cases}
\end{equation}
where $ \mathbf{W_{ep}} \in \mathbb{R}^{K \times V} $ is the trainable word embedding matrix of NPs, where each word in the vocabulary of size \textit{V} is represented as a \textit{K}-dimensional vector, and $w_{tp}^i$ is a one-hot vector indicating the location of ground truth word in the vocabulary at time step $t_p$ of phrase \textit{i}.

For a LSTM block at time step $t_p$, let $\mathbf{i}_{t_p}, \mathbf{f}_{t_p}, \mathbf{o}_{t_p}, \mathbf{c}_{t_p}$ and $\mathbf{h}_{t_p}$ denote the input gate, forget gate, output gate, memory cell and hidden state at the time step. Thus, the LSTM transition equations omitting the phrase index \textit{i} are:
\begin{equation}
\mathbf{i}_{t_p} = \sigma(\mathbf{W_i} \mathbf{x}_{t_p} + \mathbf{U_i} \mathbf{h}_{{t_p}-1} + \mathbf{b_i})~,
\end{equation}
\begin{equation}
\mathbf{f}_{t_p} = \sigma(\mathbf{W_f} \mathbf{x}_{t_p} + \mathbf{U_f} \mathbf{h}_{{t_p}-1}+ \mathbf{b_f})~,
\end{equation}
\begin{equation}
\mathbf{o}_{t_p} = \sigma(\mathbf{W_o} \mathbf{x}_{t_p} + \mathbf{U_o} \mathbf{h}_{{t_p}-1} + \mathbf{b_o})~,
\end{equation}
\begin{equation}
\mathbf{u}_{t_p} = tanh(\mathbf{W_u} \mathbf{x}_{t_p} + \mathbf{U_u} \mathbf{h}_{{t_p}-1} + \mathbf{b_u})~,
\end{equation}
\begin{equation}
\mathbf{c}_{t_p} = \mathbf{i}_{t_p} \odot \mathbf{u}_{t_p} + \mathbf{f}_{t_p} \odot \mathbf{c}_{{t_p}-1}~,
\end{equation}
\begin{equation}
\mathbf{h}_{t_p} = \mathbf{o}_{t_p} \odot tanh(\mathbf{c}_{t_p})~,
\end{equation}
\begin{equation}
\mathbf{p}_{{t_p}+1} = \text{softmax}(\mathbf{h}_{t_p})~.
\end{equation}

Here, $\sigma$ denotes the logistic sigmoid function while $\odot$ denotes elementwise multiplication. The LSTM parameters \{$\mathbf{W_i}, \mathbf{W_f}, \mathbf{W_o}, \mathbf{W_u}, \mathbf{U_i}, \mathbf{U_f}, \mathbf{U_o}, \mathbf{U_u}$\} are all matrices with dimension of $\mathbb{R}^{K \times K}$. Intuitively, each gating unit controls the extent in which information is updated, forgotten and forward-propagated while the memory cell holds the unit internal memory regarding the information processed up to current time step. The hidden state is therefore a gated, partial view of the memory cell of the unit. 

The output of the LSTM at each time step, $\mathbf{p}_{{t_p}+1} \in \mathbb{R}^V$ is equivalent to the conditional probability of word given the previous words and image, $P(w_{tp} | w_{1:{tp}-1},I)$. Its ground truth is equivalent to the input word of next time step, and an end-word token at the last time step to indicate the end of the NP. Additionally, the hidden state of last time step is used as the compositional vector representation of the NP, where 
\begin{equation}
\mathbf{z}_i = \mathbf{h}_{L_i}, \qquad \mathbf{z} \in \mathbb{R}^K~.
\end{equation} 
It is input to the AS decoder as described next.

%----------------------------------------------------------
\subsection{Abbreviated Sentence (AS) Decoder}
%----------------------------------------------------------
The AS decoder has a similar design as the phrase decoder, except the inputs, outputs and GTS, as shown in \figurename~\ref{fig:sent_RNN}. The input of the AS decoder is a complete caption describing the image, with each NP (e.g \textit{the man}) and the remaining words in the caption (e.g \textit{in}) both encoded as the input in a single time step. Let $t_s$ denotes the time step of the AS decoder and \textit{N} is the length of the caption considering each NP as a unit, the input of AS decoder $\mathbf{y_{t_s}}$ is:
\begin{equation}
\mathbf{y_{t_s}} =
\begin{cases}
\mathbf{W_{is}} \text{CNN}(I) + \mathbf{b_{is}}~, & \text{for}\ t_s=-1 \\
\mathbf{x_{ss}}~, & \text{for}\ t_s=0 \\
\begin{rcases}
\mathbf{W_{es}}w_{ts}~, & \text{if input is word} \\
\mathbf{z}_i~, & \text{if input is phrase}~i
\end{rcases}
 & \text{for}\ t_s = {1...N}~.
\end{cases}
\end{equation}

The $\mathbf{W_{is}} \in \mathbb{R}^{K \times D}$, $\mathbf{b_{is}} \in \mathbb{R}^K$, $\mathbf{x_{ss}} \in \mathbb{R}^K$ and $ \mathbf{W_{es}} \in \mathbb{R}^{K \times V} $ here are another set of trainable parameters for image embedding, start-word token and word embedding matrix of AS, while $w_{ts}$ is the one-hot vector indicator of ground truth word of time step $ts$. A new set of LSTM parameters is used for AS decoder.

Two outputs are produced by the LSTM model at each time step in the AS decoder, which are 
\begin{enumerate*}[label={\roman*)}]
\item a binary indicator that determines whether the next input is a phrase or a word (i.e. phrase indication), and 
\item a softmax prediction of the next word in the sequence of AS (i.e. word prediction).
\end{enumerate*}
The ground truth of the second output at each time step is either the last word of next phrase or the next word itself: 
\begin{equation}
\label{eq_GTS}
GTS_{t_s} = 
\begin{cases}
w_{{t_s}+1}~, & \text{if next input is word} \\
w_{L_i}^i~, & \text{if next input is phrase \textit{i}} \\
\text{end-word token}~, & \text{when } t_s = N~.
\end{cases}
\end{equation}

In our preliminary work \cite{tan2017phi}, we used a phrase token for phrase indication, which resulted in a limitation of unable to discern on the appropriateness of different NP inputs during decoding. As a compensation, a phrase selection objective was introduced to solve the limitation. However, it has a complicated training procedure, because it is optimized over multiple randomly selected NPs input at each time step when the input is a NP. To simplify the training process, we replace the phrase token and the phrase selection objective with phrase indication and softmax prediction of last word of each NP (i.e. Equation \ref{eq_GTS}, if next input is phrase \textit{i}) respectively. 

\begin{figure*}[!t]
	\centering
	\includegraphics[height=0.28\linewidth, width=0.8\linewidth]{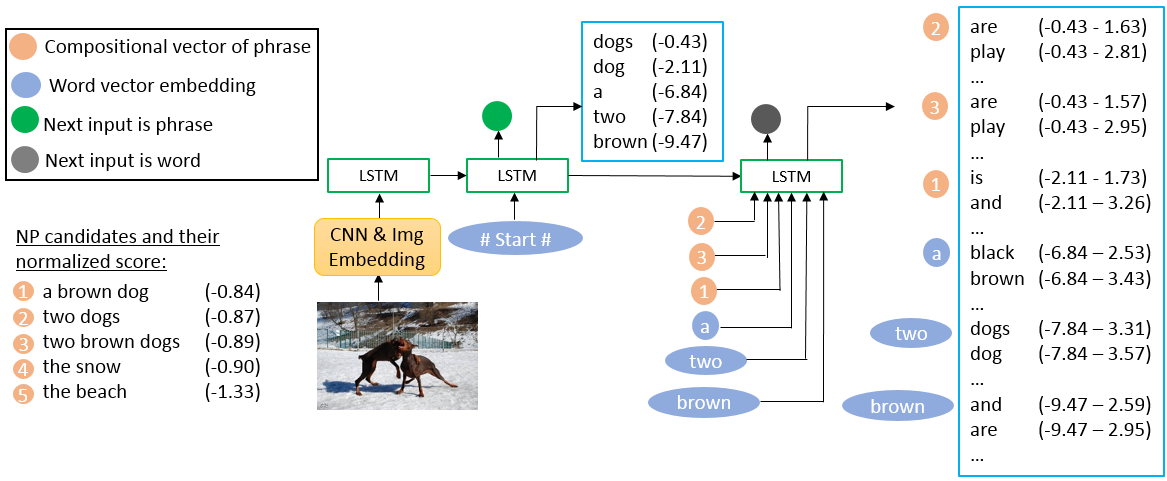}
	\caption{Example of image caption generation given a set of generated NPs ($b_s = b_p = 5$ in this example). Best viewed in color.}\vspace{-.1in}
	\label{fig:cap_gen}
\end{figure*}

%----------------------------------------------------------
\subsection{Training the phi-LSTM Model}
%----------------------------------------------------------
The objective function of our model is a log-likelihood cost function computed from the perplexity of word prediction summed with a loss from the phrase indication prediction. 

To compute the perplexity given an image \textit{I} and its description \textit{S}, let \textit{R} be the number of phrases of the sentence, while $\mathbf{p_{t}}_{p}$ and $\mathbf{p_{t}}_{s}$ be the probability output of LSTM block at time step $\mathit{t_{p}}-1$ and $\mathit{t_{s}}-1$ respectively. So, the perplexity of sentence \textit{S} conditioned on its image \textit{I} is

{\small
\begin{equation}
\log_2 \mathcal{PPL}(S|I) = -\frac{1}{M} \left[ \sum_{t_{s}=1}^{N+1} \log_2 \mathbf{p_t}_{s} + \sum_{i=1}^{R} \left[\sum_{t_{p}=1}^{L_i+1} \log_2 \mathbf{p_t}_{p} \right] \right]~,
\end{equation}
}
where $M = N + 1 + \sum_{i=1}^{R} (L_i + 1).$

We use hinge loss as the phrase indication objective to classify the next input of the AS decoder into either phrase or word. The cost function of the classifier is
\begin{equation}
\mathcal{C}_{PI} = \sum_{t_{s}=1}^{N} \kappa_{t_{s}} \sigma(1-y_{t_{s}}h_{t_{s}}\mathbf{W_{ps}})~,
\end{equation}
where $h_{t_{s}}$ is the hidden state output of the LSTM block at time step $t_{s}$, $\mathbf{W_{ps}} \in \mathbb{R}^{K \times 1}$ is trainable parameters for the classifier, while $y_{t_{s}}$ is +1 if the next input to the AS decoder is a phrase or -1 otherwise. Here, $\kappa_{t_{s}}$ scales and normalizes the objective based on the number of phrases and words in the AS. 

Hence, with \textit{P} number of training samples, the overall objective function of our model is: 
\begin{equation}
\mathcal{C(\theta)} = -\frac{1}{Q} \sum_{j=1}^{P} \left[M_j \log_2 \mathcal{PPL} (S_j|I_j) + \mathcal{C}_{PI} \right] + \lambda_\theta \cdot \parallel \theta \parallel_2^2~,
\end{equation}
where $Q = P\times\sum_{j=1}^{P} M_j~.$ It is equivalent to the average log-likelihood of word given their previous context and the image described, summed with a regularization term, $\lambda_\theta \cdot \parallel \theta \parallel_2^2$, average over the number of training samples. Here, $\theta$ is all the trainable parameters of the model. 

In summary, the proposed phi-LSTM is optimized to predict 
\begin{enumerate*}[label={\roman*)}]
	\item the next word given all previous words in each NP,
	\item the next word of AS given all previous words and phrases, and
	\item whether the next input is a phrase. 
\end{enumerate*}

%----------------------------------------------------------
\section{Image Caption Generation}
\label{sec:generation}
%----------------------------------------------------------
The phi-LSTM model generates image caption in a two-steps manner, where a list of NP candidates are first generated followed by the complete caption, both using beam search algorithm. The beam size for phrase and sentence generation are $b_p$ and $b_s$ respectively.

Generation of NPs in this work is similar to \cite{Vinyals2015}, where a given image encoded with CNN followed by a start-word token are input to the model, acting as the initial context of the phrase decoder to generate NPs. At every time step, $b_p$ words with the highest probability are sampled and input to the decoder at the next time step to infer the subsequent words. A set of $b_p$ best sequences generated up to time step $t_p$ are kept as candidates for inference of next word iteratively, until all candidates infer an end-word token. A score is then computed for each NP candidate by summing the log probability of each word normalized by the length of NP, including the end-word token:
\begin{equation}
\label{eq_gen1}
S_p = \frac{1}{L+1} \left[ \sum_{t_{p}=1}^{L+1} \log_2 \mathbf{p_t}_{p} \right]~,
\end{equation}

Among the $b_p$ NP candidates generated, at least one candidate (of highest score) is kept for each NP group that has the same last word. The remaining candidates are discarded if their score is lower than a threshold value \textit{T}, in order to improve the quality of image description formed. A total of $b_s$ complete captions are then generated from the list of NP candidates, as illustrated partially in \figurename~\ref{fig:cap_gen}. The AS decoder produces two outputs at each time step, which are next word prediction and phrase indication of next input. Thus, when the model infers that the next input is a phrase, each of the $b_s$ word candidates inferred (e.g. \textit{dogs, dog, a, two, brown} in \figurename~\ref{fig:cap_gen}) is compared with the list of NP candidates. Those NPs with last word matches the inferred words (e.g. \textit{a brown dog, two dogs, two brown dogs}) are attached to the list of beam candidates at the current time step, replacing the inferred words (e.g. beam that infers `\textit{dog}' will use NP `\textit{a brown dog}' as next input instead). Once all candidate sentences infer an end-word token, the score of each caption is computed as:
\begin{equation}
\label{eq_gen2}
S_s = -\log_2 \mathcal{PPL}(S|I) ~,
\end{equation}
where the sentence obtain the highest score is kept. %Large beam size is applied at both stages so that more candidates can be kept before normalizing the score of each sentence.

%----------------------------------------------------------
\section{Phrase Chunking, Limitations and Refinement}
\label{sec:parsing}
%----------------------------------------------------------
A quick overview on the structure of image descriptions reveals that key elements which compose the majority of captions are usually NPs that describe the dominant entities in an image, which can be either an object, group of objects or scene. These entities have equivalent abstract level as the output of our CNN encoder, and are linked with verb and prepositional phrase. Thus, NP essentially covers over half of the corpus in a language model trained to generate image description. Therefore, we partition the learning of NP and sentence structure so that they can be processed more evenly, compared to extract all phrases without considering their part of speech tag. 

This section describes 
\begin{enumerate*}[label={\roman*)}]
	\item the parsing algorithm we applied to obtain the AS-NPs pair,
	\item problems that arose from the limitation of parsing tool and our proposed algorithm, and
	\item a measure we took to reduce the influence of these errors on the training of our image captioning model. 
\end{enumerate*}

\subsection{Phrase Chunking}
To identify NPs from a training caption, we adopt the dependency parse of Stanford CoreNLP tool \cite{manning-EtAl:2014:P14-5}, which forms a structural relation tree over a sentence by providing structural relationships between words. Though it does not chunk sentence directly as in constituency parser and other chunking tools, the pattern of NP extracted is more flexible as we can select desirable structural relations. The relations we selected are:

\begin{itemize}
	\item determiner relation (\textit{det}), %links a noun phrase with its determiner, e.g. ``\textit{a man}".  
	\item numeric modifier (\textit{nummod}), %changes the meaning of noun in term of its quantity, e.g. ``\textit{two girls}".
	\item adjectival modifier (\textit{amod}), %relation modify the attribute of a noun, e.g. ``\textit{young man}".
	\item adverbial modifier (\textit{advmod}), only selected when the meaning of adjective term is modified, e.g. ``\textit{dimly lit room}", %relation changes the meaning of word that is not noun. In this work, this relation is 
	\item compound (\textit{compound}), %relates two or more nouns, where the former serves to modify the meaning of the later noun, e.g. ``\textit{basketball court}".
	\item nominal modifier for `of' \& possessive alteration (\textit{nmod:of} \& \textit{nmod:poss}), with case `of' included. %relates two nouns where the governor is possessed by the dependent, e.g. ``\textit{A man's hand}" and ``\textit{shoes of a girl}".
\end{itemize}

\begin{figure}[t]
	\includegraphics[height=0.4\linewidth, width=0.65\linewidth]{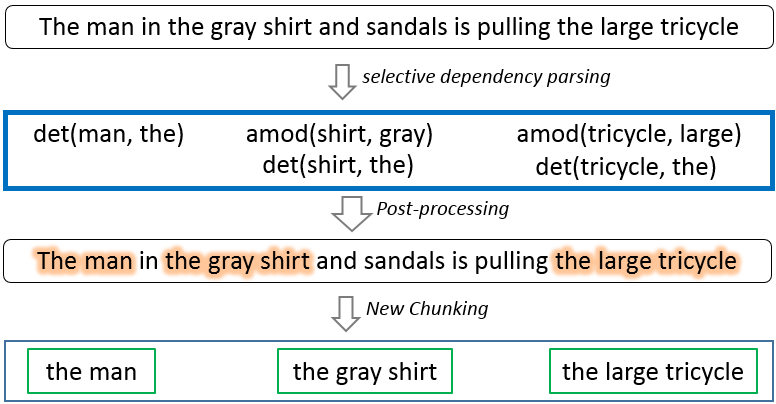}
	\centering
	\caption{An example of phrase chunking from the dependency parse.}\vspace{-.1in}
	\label{fig:phrase}
\end{figure}

The dependency parser only extracts triplet from a sentence, each made up of a governor word, a dependent word and a relation that links them, in the form of \textit{<relation (governor, dependent)>}. In order to form phrase chunks with the dependency parser, a simple post-processing step as illustrated in \figurename~\ref{fig:phrase} is carried out in this paper. That is, triplets with the same governor or dependent word which are also consecutive in the complete caption (e.g. \textit{amod(shirt, gray)} and \textit{det(shirt, the)}) are grouped as a single NP. The same applies for the consecutive triplet (e.g. \textit{det(man, the)}), while the standalone word (e.g. `in') remains as a unit in the AS.

\subsection{Limitation of Parsing Tool}
Due to the substantial ambiguity in linguistic structure, the parsing of natural language data is still an ongoing research with no perfect solution. As a result, there are always some unavoidable errors from the parser output, regardless of the chunking tool used. Asides from the dependency parser we used, we also look into constituency parser. It outputs subject and predicate of a sentence directly, and we chunk its NP constituents at the lowest level. In this section, we will compare the AS-NPs pair formed by chunking using both of the parsers\footnote{Both parsers used throughout this work are in the package of Standford CoreNLP version 3.6.0. The type of dependency parser applied is collapsed-ccprocessed-dependencies.}. The labels in the examples given below denote a complete caption (\textit{S}), AS-NPs pair formed by chunking with dependency parser (\textit{DP}), constituency parser (\textit{CP}) and dependency parser with further refinement (\textit{DP(R)}) respectively. An underlined text indicates where the AS-NP pair is wrong. %Blue and green fonts are the NPs identified by selective dependency parser before and after refinement.

One of the common errors found in the output of any parser is incorrect recognition of a verb as a noun. As a result, AS with missing object is formed, as shown in the examples below. Moreover, there are NP that does not describe entity in an image, such as `\textit{the one}' in example (c).

%\vspace{5pt}
%\resizebox{8cm}{!} {
%\begin{tabular}{l l p{8cm}}
%	\textbullet & \textit{Full}: & \textcolor{RoyalBlue}{A man} in \textcolor{RoyalBlue}{a blue shirt \underline{standing}} in \textcolor{RoyalBlue}{a garden}. \\
%	{} & \textit{DP \& CP}: & Man in \underline{standing} in garden. \\
%	\textbullet & \textit{Full}: & \textcolor{RoyalBlue}{A group of young people \underline{preparing}} to go skiing.\\
%	{} & \textit{DP}: & \underline{Preparing} to go skiing. \\
%	{} & \textit{CP}: & Group of \underline{preparing} to go skiing. \\
%\end{tabular}
%}
\vspace{3pt}
\resizebox{8cm}{!} {
\begin{tabular}{l l p{8cm}}
	(a) & \textit{S}: 	 	 & A man in a blue shirt standing in a garden.					\\
	{} 	& \textit{DP \& CP}: & \textit{a man, a blue shirt \underline{standing}, a garden}	\\
	{}	& {}				 & Man in \underline{standing} in garden. 						\\
	(b) & \textit{S}: 	 	 & A group of young people preparing to go skiing.				\\
	{} 	& \textit{DP}: 		 & \textit{a group of young people \underline{preparing}}		\\
	{}	& {}				 & \underline{Preparing} to go skiing. 							\\
	{} 	& \textit{CP}: 		 & \textit{a group, young people \underline{preparing}}			\\
	{}	& {}				 & Group of \underline{preparing} to go skiing. 				\\
	(c) & \textit{S}: 	 	 & Two men look toward the camera, while the one in front points his index finger. \\
	{} 	& \textit{DP \& CP}: & \textit{two men, the camera, the one, \underline{front points}, his index finger}	\\
	{}	& {}				 & Men look toward camera, while one in \underline{points} finger. 					\\
	(d) & \textit{S}: 	 	 & Two men and a woman on chairs outside near water. 			\\
	{} 	& \textit{DP \& CP}: & \textit{two men, a woman, \underline{near} water}			\\
	{}	& {}				 & Men and woman on chairs outside \underline{water}.			\\
\end{tabular}
}
\vspace{5pt}

From our observation, both parsers give relatively similar NP outputs. The reasons that we chose the dependency parser over the constituency parser are:
\begin{enumerate}
	\item to chunk NPs with higher constituent level, it is more intuitive to select specific dependency relation such as `\textit{nmod:of}', than specify the level of constituent NP in its parse tree.
	\item there are some cases where a past tense verb is a part of the attributes of a noun, and the dependency parser has a higher chance to recognize it as adjective. For example:
	\vspace{5pt}
	\resizebox{8cm}{!} {
	\begin{tabular}{l l p{8cm}}
		(a) & \textit{S}: 	& Two snow covered benches sit in a snow covered field.	 \\
		{} 	& \textit{DP}:	& \textit{two snow, a snow covered field}				 \\
		{}	& {}			& Snow covered benches sit in field. 					 \\
		{}	& \textit{CP}:	& \textit{two snow, a snow}								 \\
		{}	& {}			& Snow covered benches sit in snow covered field. 		 \\
		(b) & \textit{S}: 	& A red truck speeds down a tree lined street.			 \\
		{}	& \textit{DP}: 	& \textit{a red truck, a tree lined street}				 \\
		{}	& {}			& Truck speeds down street.								 \\
		{}	& \textit{CP}: 	& \textit{a red truck, a tree}							 \\
		{}	& {}			& Truck speeds down tree lined street. 					 \\
	\end{tabular}
	}
\end{enumerate}

Among all the selected dependency relations, only the nominal modifier with possessive alteration, \textit{nmod:poss} and \textit{nmod:of} parse NPs of higher constituent level. They are desired because some of the NPs chunked under these relations correspond to an entity or a group of entities within an image as we intended, as shown in examples (a) below. While there is not much controversy for \textit{nmod:poss} relation, NPs chunked from \textit{nmod:of} relation have more ambiguity on whether the whole phrase should be split into two NPs or remained as a single NP. Example (b) below shows the case where an `\textit{of}' relation is not necessary, while example (c) shows another case when the necessity of the relation is ambiguous.

\resizebox{8cm}{!} {
\begin{tabular}{l l p{8cm}}
	(a) & \textit{S}: 	& A bird washes itself in a body of water.	\\
	{} 	& \textit{DP}:	& \textit{a bird, a body of water}			\\
	{}	& {}			& Bird washes itself in water.				\\
	{}	& \textit{CP}:	& \textit{a bird, a body}					\\
	{}	& {}			& Bird washes itself in body of water. 				\\
	(b) & \textit{S}: 	& A lunch box is full of a variety of foods.		\\
	{}	& \textit{DP}:	& \textit{a lunch box, full of a variety of foods}	\\
	{}	& {}			& Box is foods.									\\
	{}	& \textit{CP}:	& \textit{a lunch box, a variety of foods}		\\
	{}	& {}			& Box is full of foods.							\\
	(c) & \textit{S}: 	& A group of men and women walk down the center of a tree-lined street.\\
	{}	& \textit{DP}:	& \textit{a group of men and women, the center of a tree-lined street}	\\
	{}	& {}			& Women walk down street. 							\\
	{}	& \textit{CP}:	& \textit{a group, the center, a tree-lined street} \\
	{}	& {}			& Group of men and women walk down center of street. \\
\end{tabular}
}

\subsection{Refinement of NPs}
\label{sec:refine}
The limitations of parser have created unwanted variations across the training data, which will in turn affect the training effectiveness of our image captioning model. In order to reduce the influences of incorrect parsing on our model, we introduce a refinement strategy between the training of our phrase decoder and the AS decoder, which will modify the AS-NPs pair based on the local statistic of the training data. That is, the phrase decoder is first trained before the overall model. Once it yields a reasonable result, a set of NPs will be generated from each of the training image. Then, the contents of each AS-NPs pair are modified based on the generated NPs, by gradually restoring the non-inferred first word into its AS, followed by the non-inferred last word. The details of our refinement algorithm is shown in Algorithm \ref{alg_refinement}, where \textit{K} is a chunked NP start with word $W_s$ and end with word $W_e$, while $G_s$ and $G_e$ are a set of first words and last words of the generated NPs respectively.

\begin{algorithm}
\KwData{$K, W_s, W_e, G_s,G_e$}
\KwResult{AS, K}
\tcc{refinement step 1}
\While{$W_s \not\subset G_{s}$ \upshape{\textbf{and}} |K| > 0}{
	remove $W_s$ from K and update $W_s$\;
}
\uIf{|K|=1}{restore K completely into AS\;}
\ElseIf{|K|>1}{restore the removed words from original \textit{K} into AS.}
\tcc{refinement step 2}
\While{$W_e \not\subset G_{e}$ \upshape{\textbf{and}} |K| > 0}{
	remove $W_e$ from K and update $W_e$\;
}
\uIf{|K|<2}{restore \textit{K} completely into AS\;}
\Else{restore the removed words from original \textit{K} into AS.}
\caption{NP refinement algorithm.}
\label{alg_refinement}
\end{algorithm}

%\begin{enumerate}
%\item compare $W_s$ with the first word of all generated NPs, $G_s$s, and removed $W_s$ from \textit{K} iteratively until there is a match between $W_s$ and $G_s$s. Keep \textit{K} as it is if none of the words within \textit{K} match with $G_s$s, but restored \textit{K} completely into its AS if there is only one word remained in \textit{K} (this usually happens when the noun in \textit{K} is an out-of-vocab word). Otherwise, restored the removed words from original \textit{K} into AS.
%\label{item:refine1}
%\item compare $W_e$ with the last word of all generated NPs, $G_e$s. If none of the generated NPs end with $W_e$, restore the last word iteratively to its abbreviated sentence until $W_e$ matches with any of the $G_e$s. \textit{K} will be completely restored to its AS if only the first word matches with any of the $G_e$s, and if all words that made up \textit{K} are not inferred as last word by the trained phrase decoder.
%\label{item:refine2}
%\end{enumerate}

The examples below show the difference between the AS formed from our proposed phrase chunking approach described earlier before and after refinement. Example (a) shows where refinement step 1 comes into play, as none of the generated NPs start with word `\textit{full}' while some of them start with word `\textit{a}'. Example (b) is fixed with refinement step 2, as word `\textit{standing}' is not inferred as last word of all generated NPs. In example (c), phrases \textit{the one}, \textit{front points} and \textit{his index finger} are restored to its AS, because our phrase decoder which uses image alone as its context is incapable of generating NPs end with word `\textit{one}', `\textit{points}' and `\textit{finger}'. These three phrases do not correspond to any dominant entities within the image, and thus seldom occur among captions of similar images. In fact, `\textit{the one}' cannot be generated from the image content alone, as it needs its subject (`\textit{two men}') as previous context. On the other hand, word `\textit{camera}' is inferred due to the statistic of training data, as there are a lot of captions end with '\textit{looking at the camera}' for images showing the frontal view of human. Example (d) shows the case where our trained phrase decoder automatically decides which entity to be kept based on the statistic of the training data.

\vspace{3pt}
\resizebox{8cm}{!} {
\begin{tabular}{l l p{8cm}}
	(a) & \textit{S}:	  & A lunch box is full of a variety of foods.		\\
	{}	& \textit{DP}:	  & \textit{a lunch box, full of a variety of foods}\\
	{}	& {}			  & Box is foods.									\\
	{}	& \textit{DP(R)}: & \textit{a lunch box, a variety of foods}		\\
	{}	& {}			  & Box is full of foods.							\\
	(b) & \textit{S}:	  & A man in a blue shirt standing in a garden.		\\
	{}	& \textit{DP}:	  & \textit{a man, a blue shirt standing, a garden}	\\
	{}	& {}			  & Man in standing in garden.						\\
	{}	& \textit{DP(R)}: & \textit{a man, a blue shirt, a garden}			\\
	{}	& {}			  & Man in shirt standing in garden.				\\
	(c) & \textit{S}:	  & Two men look toward the camera, while the one in front points his index finger. \\
	{}	& \textit{DP}:	  & \textit{two men, the camera, the one, front points, his index finger}			\\
	{}	& {}			  & Men look toward camera, while one in points finger.								\\
	{}	& \textit{DP(R)}: & \textit{two men, the camera}													\\
	{}	& {}			  & Men look toward camera, while the one in front points his index finger.			\\
	(d) & \textit{S}:  	  & A group of men and women walk down the center of a tree-lined street.	\\
	{}	& \textit{DP}:	  & \textit{a group of men and women, the center of a tree-lined street}	\\
	{}	& {}			  & Women walk down street.													\\
	{}	& \textit{DP(R)}: & \textit{a group of men, the center of a tree-lined street}				\\
	{}	& {}			  & Men and women walk down street. 										\\
\end{tabular}
}

\section{Experiment}
\subsection{Datasets}
The proposed phi-LSTM model is tested on three benchmark datasets - Flickr8k \cite{rashtchian2010collecting}, Flickr30k \cite{young2014image}, and MS-COCO \cite{lin2014microsoft}. These datasets consist of 8000, 31000 and 123287 images respectively. Each image is annotated with at least five image descriptions prepared by human from crowd sourcing. We follow the publicly available dataset splits\footnote{http://cs.stanford.edu/people/karpathy/deepimagesent/} used in \cite{karpathy2015deep}. That is, the validation and testing set each contains 1000 images for Flickr8k \& Flickr30k datasets, and 5000 images for MS-COCO dataset. The rest of the images are used for training. 

\subsection{Evaluation Metrics}
We employ five automatic metrics, including BiLingual Evaluation Understudy (BLEU) \cite{papineni2002bleu}, Recall-Oriented Understudy for Gisting Evaluation (ROUGE) \cite{lin2004rouge}, Metric for Evaluation of Translation with Explicit ORdering (METEOR) \cite{banerjee2005meteor},  Consensus-based Image Description Evaluation (CIDEr) \cite{vedantam2015cider} and Semantic Propositional Image Caption Evaluation (SPICE)\cite{anderson2016spice} to evaluate the quality of the generated image captions. BLEU metric measures the precision of \textit{n}-grams matching between a generated caption and all reference sentences, while ROUGE metric measures the recall instead of precision. Here, we only reported ROUGE-L which uses the longest common sequence instead of n-grams. METEOR aligns generated caption and reference string by mapping each unigram using three different modules, which are ``exact'', ``porter stem'' and ``WordNet synonymy'' modules. The final score is the F-mean computed from the number of unigram mapping. CIDEr metric combines the average cosine similarity of each \textit{n}-gram between the generated caption and references. It gives lower weight to \textit{n}-grams that commonly occur across all reference captions in the dataset. Lastly, SPICE metric parses image caption and its references into a scene graph to form tuples for each semantic proposition. Then, it computes the F-score defined over the conjunction of all logical tuples.

\vspace{-3pt}
\subsection{Experimental Details}
Aside from our proposed phi-LSTM model, we have conducted experiment on a baseline model which process image caption as a sequence of words. It is basically a reimplementation of work described in \cite{Vinyals2015}, but without ensembling multiple trained models and using VGGnet \cite{simonyan2014very} instead of GoogleLeNet \cite{szegedy2015going} to encode image for fair comparison with our model. All experimental settings in the baseline model and ours are the same unless stated otherwise. 

During the training stage, we use raw caption without any preprocessing as input to the language parser in order to get a more appropriate AS-NPs pair. Then, all words in the AS-NPs pair are converted to lower case, with some punctuations removed, and word that occurs less than 5 times in the training data discarded, so that the tokenization of our image captions are consistent with that of \cite{karpathy2015deep}. To avoid gradient explosion due to overlength caption (relative to average length of all training data), we truncate sentence as specified in Table \ref{trunc}. For the overlength NPs, we truncate the first few words instead of last few words, because the latter part of NPs usually hold more significant semantic content. The length of the AS-NPs pair considered are those after the refinement step described in Section \ref{sec:refine}. The truncate length is decided such that the number of captions affected are less than 0.5\% of the whole training data.  

\begin{table}[t]
	\renewcommand{\arraystretch}{1.3}
	\caption{Caption truncation setting.}
	\label{trunc}
	\begin{tabular}{l l c c}
		\toprule
		\textbf{Dataset}	& \textbf{Model}	& \textbf{Truncate length}	& \textbf{Captions affected}	\\
		\hline
		Flickr8k			& Baseline			& 24						& 0.25\%						\\
		{}					& phi-LSTM (AS)		& 20						& 0.24\%						\\
		{}					& phi-LSTM (NP)		& 7							& 0.12\%						\\
		\hline
		Flickr30k			& Baseline			& 36						& 0.25\%						\\
		{}					& phi-LSTM (AS)		& 30						& 0.29\%						\\
		{}					& phi-LSTM (NP)		& 7							& 0.12\%						\\
		\hline
		MS-COCO				& Baseline			& 23						& 0.26\%						\\
		{}					& phi-LSTM (AS)		& 18						& 0.35\%						\\
		{}					& phi-LSTM (NP)		& 7							& 0.36\%						\\
		\bottomrule
	\end{tabular}
\end{table}

The CNN encoder we use is the VGG-16 \cite{simonyan2014very} pre-trained on ImageNet \cite{deng2009imagenet} classification task, but without fine-tuning the CNN parameters. The LSTM decoder with hidden size of \textit{K}=256 (Flickr8k) and \textit{K}=512 (Flickr30k \& MS-COCO) is employed. Our model is optimized with RMSprop \cite{rmsprop}, using a minibatch of 300(Flickr8k), 500(Flickr30k) and 700(MS-COCO) image-sentence pair per iteration. The learning rate is set to 0.001, and dropout regularization \cite{srivastava2014dropout} is employed to avoid overfitting. %After the refinement process described in Section \ref{sec:refine}, the AS decoder is trained independently without updating the parameters of phrase decoder. Then, the overall decoder is fine-tuned with a lower learning rate. %We also found that constantly updating the word embedding matrices, $\mathbf{W_{ep}}$ and $\mathbf{W_{es}}$ causes over-fitting (BLEU-scores of generated captions reduced). Therefore, $\mathbf{W_{ep}}$ is only updated during the first five epochs when training the phrase decoder and overall model, while $\mathbf{W_{es}}$ is only updated for five epochs when training of overall model.

During the testing stage, we found that our proposed model generates better caption with large beam size while the baseline model works better with small beam size due to overfitting, as stated in \cite{vinyals2017show}. Thus, we compare our model using beam size of $b_p$=30 and $b_s$=20, with the baseline model tested with beam size of \textit{b}=3 and \textit{b}=20.

\begin{figure*}[t]
	\centering
	\includegraphics[height=0.15\linewidth, width=0.8\linewidth]{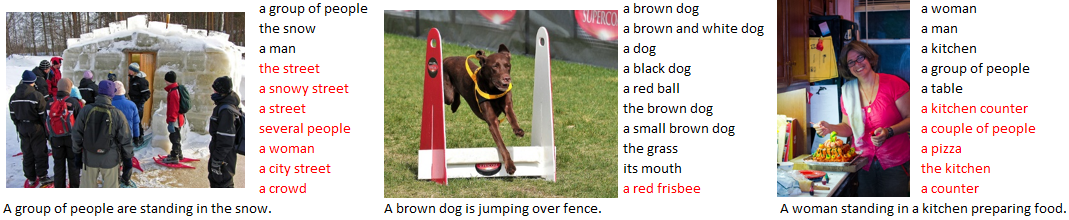}
	\caption{Examples of NPs generated from image. Red fonts indicate that the score of NP, $S_p$ is lower than threshold \textit{T}. Complete caption generated from the NP candidates are shown at the bottom of each image.}\vspace{-2pt}
	\label{fig:phrase_eg}
\end{figure*}

%\begin{figure*}[!t]
%	\centering
%	\subfloat[]
%	{
%		\includegraphics[width=0.24\linewidth]{thres8k}
%	}
%	\subfloat[]
%	{
%		\includegraphics[width=0.24\linewidth]{thres30k}
%	}
%	\subfloat[]
%	{
%		\includegraphics[height=0.43\linewidth, width=0.48\linewidth]{threscoco}
%	}
%	\caption{Effect of threshold \textit{T} on different metrics and number of unique sentence captions generated in (a) Flickr8k, (b) Flickr30k and (c) MS-COCO datasets.}
%	\label{fig:thres}
%\end{figure*}
\begin{figure*}[!t]
	\centering
	\includegraphics[height=0.3\linewidth, width=0.8\linewidth]{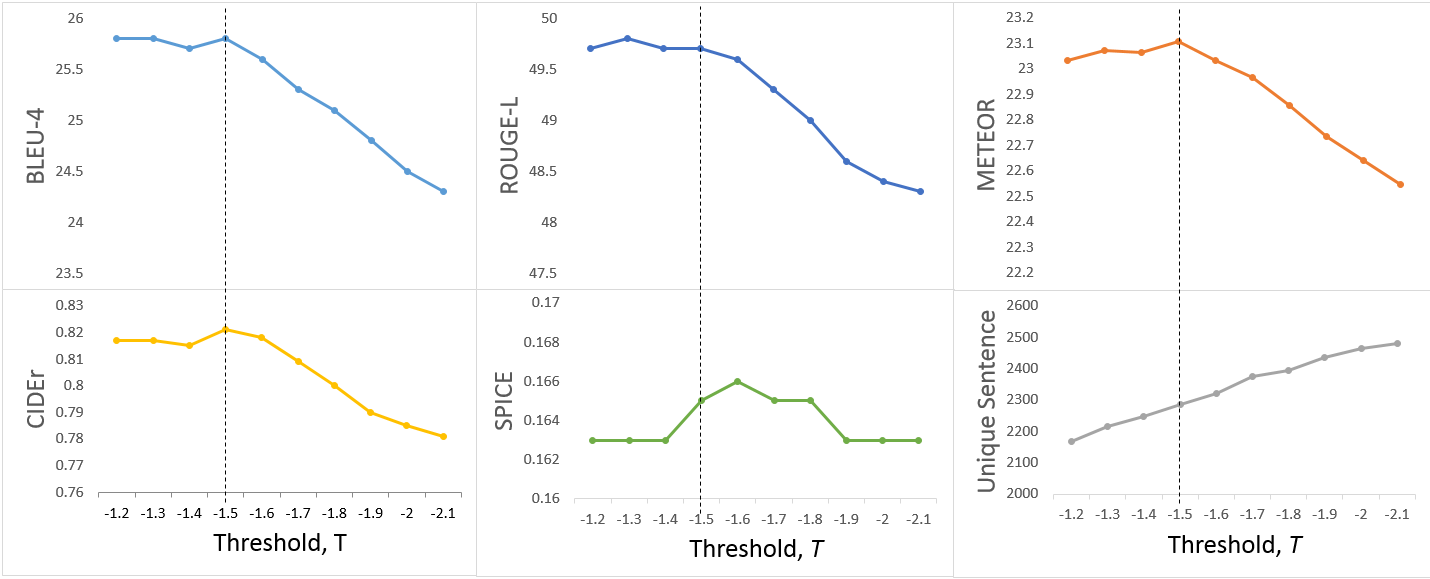}
	\caption{Effect of threshold \textit{T} on different metrics and number of unique captions generated in MS-COCO dataset.}\vspace{-2pt}
	\label{fig:thres}
\end{figure*}

When multiple NP candidates with the same last word are generated, we only keep the candidates with score higher than a predefined threshold \textit{T} for complete caption generation. Some examples of the generated NPs are shown in \figurename~\ref{fig:phrase_eg}. To choose an appropriate value of \textit{T}, we examine the changes of several metrics and sentence uniqueness on the generated captions using varying threshold value \textit{T} for each dataset, with example of MS-COCO dataset shown in \figurename~\ref{fig:thres}. It is observed that all the \textit{n}-grams metrics (BLEU, CIDEr, METEOR and ROUGE-L) gradually increase with the threshold, and reach optimum at \textit{T}=-1.6 for Flickr8k and Flickr30k datasets, and \textit{T}=-1.5 for MS-COCO dataset. Further increment of \textit{T} yields different effect on different \textit{n}-grams metrics, where BLEU and CIDEr decrease while METEOR and ROUGE-L fluctuate irregularly. Besides that, the sentence uniqueness constantly reduces with the increment of \textit{T} as a result of less choice of NP candidates. We also notice that there are not much changes in the SPICE metric, where the score fluctuate within the range of 0.163 - 0.165 across varying value of \textit{T} in the MS-COCO dataset. This shows that the threshold value \textit{T} only affects words' order and does not help much in predicting the correct objects, attributes and relations. 

\subsection{Comparison with State-of-the-Art Models}
Table \ref{result} shows the performance of our proposed model in comparison with the current state-of-the-art models. When compare with the methods that use only the CNN as encoder, our model performs better or comparable to all other state-of-the-art models, including the phrase-based models proposed by Lebret et al. \cite{lebret2015phrase} and Ushiku et al. \cite{ushiku2015common}. Note that our current model has a lower BLEU-1 and BLEU-2 score but a higher BLEU-3 and BLEU-4 score when compared to our preliminary results published in \cite{tan2017phi}. This is because we have added the length normalization in our beam search algorithm (Equations \ref{eq_gen1}-\ref{eq_gen2}) in order to generate longer caption. As reported in \cite{jia2015guiding}, lower order of the BLEU metrics is bias towards short sentence, especially when the brevity penalty is set to 1 (i.e. without brevity penalty)\footnote{Most authors of the SOTA models in Table \ref{result} did not report about the brevity penalty they set for BLEU evaluation. Nevertheless, the default setting of publicly available code in https://github.com/karpathy/neuraltalk and https://github.com/tylin/coco-caption are both without brevity penalty. Thus, we assume that this is the setting others used.}. Thus, we increase the average length of our generated caption for a better comparison with other models. On top of that, we have added a NPs refinement strategy (Algorithm \ref{alg_refinement}) and replace our phrase selection objective with predicting last word of each NP using softmax in AS decoder (Equation \ref{eq_GTS}, if next input is phrase \textit{i}). We observe approximately 1 BLEU score gained from the phrase refinement algorithm. 

Compared to Skel-Key model \cite{wang2017skeleton} which has a very similar architecture to ours, their performance is much better mainly due to three components. First, they employed a better image model, the ResNet-200 while we only use VGG16. In order to cope with the large dimension of the fully convolutional ResNet model, they have to set their LSTM hidden layer dimension to as large as 1800 for Skel-LSTM (sentence level) and 1024 for Attr-LSTM (phrase level), while we only set to 512 at both level for MS-COCO dataset. Secondly, they fine-tuned their image model while we fix all the CNN parameters during training. Thirdly, they implemented attention mechanism to generate image caption, while we do not. Since the objective of our work is to investigate the capability of a phrase-based image captioning model, as compared to a similar model trained on flat sequences, we do not implement attention mechanism or provide extra information to our model, as it is beyond the scope of this paper. Nevertheless, we argue that our model is comparable to the soft-attention model \cite{xu2015show}. The attention mechanism requires extra computation of relative importance of each location in feature maps at every time step. As for Review Network \cite{yang2016review}, the relative importance of each location in feature maps is computed during a total of eight review time steps using LSTM, and all outputs of review LSTM are attended during the decoding of image caption. Finally, one interesting finding from the Skel-Key model is that there is no performance gain\footnote{This result is before the removal of word `\textit{a}'} between Skel-Key model and their baseline model (i.e without the skeleton-attribute decomposition) trained on complete captions in the BLEU, ROUGE-L, METEOR and CIDEr metrics. On the contrary, our model outperforms our baseline in all three datasets. %We believe that this is the side-effect caused by the imperfection of parser that creates noise across the training data. We manage to reduce this noise by introducing a refinement step described in Section \ref{sec:refine}, and thus our results are consistently better than baseline in all datasets. Nevertheless, their finding of improvement gained over baseline in SPICE metric is consistent with ours, as shown in Table \ref{spice}.

\begin{table*}[t]
	\renewcommand{\arraystretch}{1.3}
	\caption{Performance of phi-LSTM and other state-of-the-art methods evaluated with automatic metrics. B-\textit{n}, MT, RG and CD stands for \textit{n}-gram BLEU, METEOR, ROUGE-L and CIDEr respectively. $^\dagger$ indicates that the results is obtained by ensembling multiple trained models, while (w.o.r) and (w.r) refer to with and without phrase refinement respectively}
	\label{result}
	\centering
	\begin{tabular*}{\linewidth}{l | c c c c c | c c c c c | c c c c c c c }
		\toprule
		{} & \multicolumn{5}{c|}{\textbf{Flickr8k}} & \multicolumn{5}{c|}{\textbf{Flickr30k}} & \multicolumn{7}{c}{\textbf{MS-COCO}} \\
		\hline
		Models 							 & B-1	& B-2  & B-3  & B-4	 & MT	& B-1  & B-2  & B-3  & B-4  	& MT	 & B-1  	& B-2 		& B-3  & B-4  & MT	 & RG 	& CD	\\
		\hline
		mRNN \cite{mao2014deep} 	   	 & -	& -    & - 	  & -    & -    & 60.- & 41.- & 28.- & \textbf{19.-} & - & 67.- 	& 48.-		& 35.- & 25.- & - 	 & - 	& -		\\
		DeepVS \cite{karpathy2015deep}	 & 57.9 & 38.3 & 24.5 & 16.0 & 16.7 & 57.3 & 36.9 & 24.0 & 15.7 	& 15.3	 & 62.5 	& 45.0 		& 32.1 & 23.0 & 19.5 & -	& 0.66	\\
		LRCNN \cite{donahue2015long}	 & -	& -    & - 	  & -    & -    & 58.7 & 39.1 & 25.1 & 16.5 	& -	 	 & 66.9 	& 48.9 		& 34.9 & 24.9 & -	 & -	& -		\\
NIC \cite{Vinyals2015}$^\dagger$\footnote& 63.- & 41.- & 27.- & -    & -    & 66.3 & 42.3 & 27.7 & 18.3 	& -		 & 66.6 	& 46.1 		& 32.9 & 24.6 & -	 & -	& -		\\
		PbIC \cite{lebret2015phrase}	 & -	& -    & - 	  & -    & -    & 59.- & 35.- & 20.- & 12.- 	& -		 & \textbf{70.-} & 46.- & 30.- & 20.- & -	 & -	& -		\\
		CoSMos \cite{ushiku2015common}	 & -	& -    & - 	  & -    & -    & -	   & -    & - 	 & -    	& -      & 65.-	& \textbf{49.-}	& 32.- & 20.- & 20.- & -	& -		\\
		phi-LSTM \cite{tan2017phi}		 & \textbf{63.6} & 43.6 & 27.6 & 16.6 & -  & \textbf{66.6} & \textbf{45.8} & 28.2 & 17.0 & - & - & -& -    & -    & -    & -    & - 	\\
		Baseline, \textit{b}=3			 & 57.6	& 39.2 & 26.1 &	17.5 & 19.1 & 57.0 & 38.5 & 25.9 & 17.3 	& 17.3 	 & 65.2 	& 47.5 		& 34.3 & 25.2 & 22.6 & 49.3 & 0.78	\\
		Baseline, \textit{b}=20			 & 56.2	& 38.0 & 25.3 &	16.7 & 19.0 & 57.0 & 38.3 & 25.7 & 17.3 	& 17.8 	 & 61.7 	& 43.7 		& 31.4 & 23.1 & 22.4 & 47.7 & 0.72	\\
		phi-LSTMv2 (w.o.r)				 & 61.5	& 43.1 & 29.6 &	19.7 & 19.9 & 60.6 & 41.2 & 27.8 & 18.6 	& 18.1 	 & - 		& - 		& - & - & - & - & -	\\
		phi-LSTMv2 (w.r)				 & 62.7 & \textbf{44.4} & \textbf{30.7} & \textbf{20.8} & \textbf{20.2} 
		& 61.5 & 42.1 & \textbf{28.6} & \textbf{19.3} & \textbf{18.2} 
		& 66.6 & 48.9 & \textbf{35.5} & \textbf{25.8} & \textbf{23.1} & \textbf{49.7} & \textbf{0.82} \\
		\hline
		\multicolumn{18}{l}{\textit{State-of-the-art results using attention mechanism}} \\
		\hline
		Soft-Atten\cite{xu2015show}		 & 67.0 & 44.8 & 29.9 & 19.5 & 18.9 & 66.7 & 43.4 & 28.8 & 19.1 & 18.5 & 70.7 & 49.2 & 34.4 & 24.3 & 23.9 & -	 & -	\\
		Hard-Atten\cite{xu2015show}		 & 67.0 & 45.7 & 31.4 & 21.3 & 20.3 & 66.9 & 43.9 & 29.6 & 19.9 & 18.5 & 71.8 & 50.4 & 35.7 & 25.0 & 23.0 & -	 & -	\\
		Review \cite{yang2016review}	 & -	& -    & - 	  & -    & -    & -	   & -    & - 	 & -    & -    & -	  & -	 & -	& 29.- & 23.7 &	-	 & 0.88 \\
		Skel-Key \cite{wang2017skeleton} & -	& -    & - 	  & -    & -    & -	   & -    & - 	 & -    & -    & 74.2 & 57.7 & 44.0 & 33.6 & 26.8 & 55.2 & 1.07 \\
		\hline
		\multicolumn{18}{l}{\textit{State-of-the-art results using extra information / extra information+attention mechanism}} \\
		\hline
		g-LSTM \cite{jia2015guiding}	& 64.7	& 45.9 & 31.8 & 21.6 & 20.2	& 64.6 & 44.6 &	30.5 & 20.6	& 17.9 & 67.0 &	49.1 & 35.8	& 26.4 & 22.7 &	- 	 & 0.81	\\
		ACVT \cite{wu2017image} 		& 74.-	& 54.- & 38.- &	27.- & -	& 73.- & 55.- & 40.- & 28.- & -	   & 74.- & 56.- & 42.-	& 31.- & 26.- &	-	 & 0.94 \\
Reg-Atten\cite{fu2016aligning}$^\dagger$& 63.9	& 45.9 & 31.9 & 21.7 & 20.4	& 64.9 & 46.2 & 32.4 & 22.4	& 19.4 & 72.4 &	55.5 & 41.8	& 31.3 & 24.8 & 53.2 & 0.96 \\
Sem-Atten\cite{you2016image}$^\dagger$	& -		& -    & - 	  & -    & -    & 64.7 & 46.0 & 32.4 & 23.0	& 18.9 & 70.9 & 53.7 & 40.2	& 30.4 & 24.3 &	-	 & -	\\
		\bottomrule
	\end{tabular*}
\end{table*}

\begin{table*}[t]
	\renewcommand{\arraystretch}{1.3}
	\caption{Performance of phi-LSTM and baseline model evaluated with SPICE measurements on MS-COCO dataset.}
	\label{spice}
	\centering
	\begin{tabular*}{0.78\linewidth}{c| c c c c c c c c c}
		\toprule
		Models						& SPICE	& Precision	& Recall	& Object	& Relation	& Attribute	& Size	& Color	& Cardinality\\
		\hline
		Baseline, \textit{b}=3		& 0.154			 & 0.395		  & 0.098		   & 0.293			& \textbf{0.039} & 0.059		  & 0.025		   & 0.062			& \textbf{0.005} \\
		Baseline, \textit{b}=20		& 0.150			 & 0.386		  & 0.095		   & 0.284			& 0.033			 & 0.064		  & 0.023		   & 0.070			& 0.000			 \\	
		phi-LSTM					& \textbf{0.165} & \textbf{0.449} & \textbf{0.104} & \textbf{0.310}	& 0.038 		 & \textbf{0.076} & \textbf{0.036} & \textbf{0.100} & 0.002			 \\
		\bottomrule
	\end{tabular*}
\end{table*}

Other state-of-the-art methods that outperform ours require extra information as input to their model on top of the CNN-encoded feature. For instance, g-LSTM\cite{jia2015guiding} provides semantic representation of cross-modal retrieval model as extra input to their LSTM model, while ACVT model \cite{wu2017image} requires training of external module to convert image into attributes as input to their decoder. Region attention \cite{fu2016aligning} requires extraction of image regions, training of ``objectness'' classifier for each region and computation of relative importance of image regions at every time step. Their performance is further boosted by using a better image model (ResNet-152) with additional scene-specific context and model ensembling.  Lastly, semantic attention model \cite{you2016image} requires both the training of attributes detector and the computation of relative importance of each attribute at every time step.

\section{Analysis of phi-LSTM model in Comparison to Its Sequence Model Counterpart}

\subsection{SPICE Metric Evaluation}
From the evaluation of SPICE metrics shown in Table \ref{spice}, we observe that there are improvements in terms of object, attribute, size, and color by decoding image caption in a phrase-based hierarchical manner. All these improvements gained are at the object level. This is because we have essentially broken down the generation process of subsequences from global sequence with our proposed model. Therefore, the phrase decoder does not need to shift the time-scale of generative process repeatedly, and can focuses on a particular aspect of image when generating the NPs. Note that this is difference from the attention mechanism implemented in \cite{xu2015show,yang2016review,fu2016aligning,you2016image}, which provides a guidance to transit attention to image region in a sequence that spreads out over multiple time-scales. Our model does not attend to image regions, but fixes the time-scale of subsequence decoder at the object level. Nonetheless, our model has a global sequence of mixed time-scales, as non-object phrases are decoded in multiple time steps at the sentence level. 

%\addtocounter{footnote}{-1}
\footnotetext{The score reported here is cited from \cite{xu2015show}, in which the authors claimed that they obtained the missing metrics from authors of \cite{Vinyals2015} through personnel communications.} 

There are no improvement in terms of the object relations, as the CNN encoder we used does not hold any information regarding to the relative position of the objects. Therefore, object relations are mostly inferred from the local statistic of training data and posture of subject in the image. Lastly, cardinality is a measurement of correctness in terms of object counting. The low score obtained by both the baseline and our model indicate that neither are able to count objects in image, as the CNN encoder are trained for object recognition instead of counting. Nevertheless, there are still small chances to guess object counts correctly, when the interaction of multiple subjects of same class (\textit{e.g. men}) is captured in the image.

\begin{table}
\renewcommand{\arraystretch}{1.3}
\caption{Measure of caption uniqueness and novelty. A higher `seen' percentage indicates that the generated captions are less novel. The number of unique words of all captions is shown under `Words', where `Within vocab.' considers only words that are in the training corpus. }
\label{cap_anal}
\centering
\begin{tabular}{l| c| c| c| c| c}
	\toprule
	\multirow{2}{*}{\textbf{Models}}	& \multicolumn{3}{c|}{\textbf{Sentence}}& \multicolumn{2}{c}{\textbf{Words}}\\
	\cline{2-6}
	{}									& Unique	& Seen		& Avg. 			& 	Actual		& Within			\\
	{}									& {}		& {}		& length		& {}			& vocab.				\\
	\hline
	\multicolumn{6}{l}{\textit{Flickr8k}}																			\\
	\hline
	References							& 99.84\%	& 1.20\%	& 10.87			&	3147		& 1919				\\
	Baseline (\textit{b}=3)				& 58.70\%	& 10.80\%	& 11.06			& 	-			& 196				\\
	Baseline (\textit{b}=20)			& 54.40\% 	& 12.20\%	& 11.54			&	-			& 201				\\
	phi-LSTM							& 67.70\%	& 7.40\%	& 9.72 			& 	-			& 212				\\
	\hline
	\multicolumn{6}{l}{\textit{Flickr30k}}																			\\
	\hline
	References							& 99.96\%	& 0.30\%	& 12.39			&	4204		& 3561				\\
	Baseline (\textit{b}=3)				& 65.70\%	& 10.70\%	& 12.40			& 	-			& 348				\\
	Baseline (\textit{b}=20)			& 58.90\% 	& 9.40\%	& 12.81			&	-			& 328				\\
	phi-LSTM							& 77.20\%	& 9.30\%	& 11.07			& 	-			& 375				\\
	\hline
	\multicolumn{6}{l}{\textit{MS-COCO}}																			\\
	\hline
	References							& 99.22\%	& 5.56\%	& 10.44			&	7241		& 5949				\\
	Baseline (\textit{b}=3)				& 38.06\%	& 63.54\%	& 10.12			& 	-			& 517				\\
	Baseline (\textit{b}=20)			& 24.54\% 	& 77.32\%	& 10.60			&	-			& 457				\\
	phi-LSTM							& 46.42\%	& 48.54\%	& 9.81			& 	-			& 548				\\
	\bottomrule
\end{tabular}
\end{table}

\subsection{Evaluation on Uniqueness and Novelty of Caption}
It has been pointed out that multimodal RNN-based approach tends to reconstruct previously seen captions\cite{devlin2015language}. Hence, we compare our model with baseline in terms of the uniqueness and novelty of the generated captions. We compute and tabulate  
\begin{enumerate*}[label={\roman*)}]
	\item the percentage of unique captions generated,
	\item the percentage of generated captions that are seen in the training data,
	\item the average length of the captions, and
	\item the number of unique words generated
\end{enumerate*}
in Table \ref{cap_anal}. To obtain an upper bound of performance under these measures, we evaluate the five human annotated captions of the set of same test images as reference. 

From Table \ref{cap_anal}, we observe that our model can generate more unique and novel (not seen in training data) captions, when compared with the baseline in all three datasets. Although the average length of our captions is shorter than the baseline, it is only about one word less when compared with the human annotated captions. In our experiment, the vocabulary size of Flickr8k, Flickr30k and MS-COCO datasets are 2538, 7413 and 9996 words respectively. Therefore, there are a total of 1228, 643 and 1292 out-of-vocabulary words in the test set of the three datasets respectively, which would penalize all the automatic metrics we used. Assume that all within-vocabulary words in the reference captions are the upper bound of test image relevant words a well-trained image captioning model can infer, we observe that both our model and baseline can only generate captions that made up of around 10\% of all possible words. Nevertheless, the number of unique words generated using our model is still higher than the baseline which has a longer average caption length.

\begin{table*}
	\renewcommand{\arraystretch}{1.3}
	\caption{Top-5 least seen words that are inferred in the generated captions. Highlighted words are used correctly in describing the image content.}
	\label{least_seen}
	\centering
	\begin{tabular}{c c| c c| c c| c c| c c| c c}
		\toprule
		\multicolumn{4}{c|}{\textbf{Flickr8k}}					  & \multicolumn{4}{c|}{\textbf{Flickr30k}}					  & \multicolumn{4}{c}{\textbf{MS-COCO}}						\\
		\hline
		\multicolumn{2}{c|}{Ours} & \multicolumn{2}{c|}{Baseline} & \multicolumn{2}{c|}{Ours} & \multicolumn{2}{c|}{Baseline} & \multicolumn{2}{c|}{Ours} & \multicolumn{2}{c}{Baseline}	\\
		\hline
		Words		& Seen		  & Words			& Seen		  & Words			& Seen	  & Words			& Seen		  & Words			& Seen	  & Words					& Seen	\\
		\hline
		bubble		& 34		  & stage			& 39		  & \colorbox{green}{tackled} & 48  & tablecloth& 40	 	  & clearly			& 70	  & headboard				& 117	\\
		\colorbox{green}{kayak}	& 54	& log		& 42		  & cows 			& 49	  & tackled			& 48		  & \colorbox{green}{unripe}	& 94	& drivers		& 183	\\
		\colorbox{green}{driving}		& 55	& \colorbox{green}{snowboarding}	& 44		  
		& \colorbox{green}{chalkboard}	& 52	& \colorbox{green}{dune}			& 82			  
		& \colorbox{green}{printer}		& 123	& \colorbox{green}{racquets}		& 184	\\
		tent		& 57		  & hind			& 44	& \colorbox{green}{tackle}	& 86  & \colorbox{green}{formations}	& 82  & hangar	& 134	  & \colorbox{green}{backs}	& 219	\\
		book		& 61		  & \colorbox{green}{kayak}	& 54  & handstand		& 91	  & fruits	& 88	& \colorbox{green}{towering}	& 176	& \colorbox{green}{herself}	& 237	\\
		\bottomrule
	\end{tabular}
\end{table*}

\begin{figure*}[!t]
	\centering
	\includegraphics[height=0.8\linewidth, width=\linewidth]{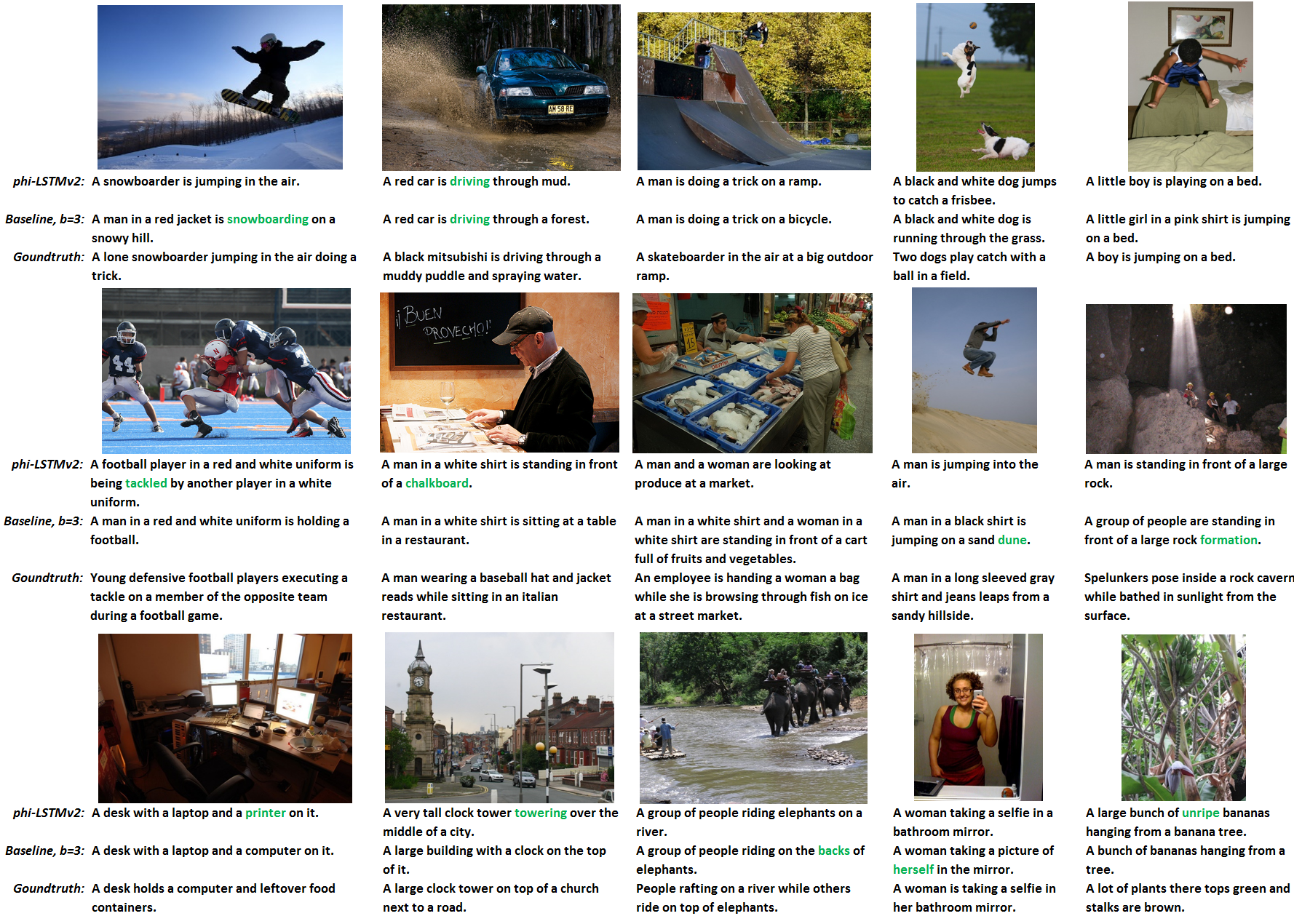}
	\caption{Examples of caption generated in Flickr8k (1st row), Flickr30k (2nd row) and MS-COCO (3rd row) datasets. The least seen words that are used correctly in the description are in green.}
	\label{fig:caption_eg}
\end{figure*}

\subsection{Model Limitations Observed with Qualitative Analysis}
To get further insights on how the number of occurrence of each word in the training corpus affects the word prediction when generating caption, we record the top five, {\it least seen} words that are inferred by both models, as shown in Table \ref{least_seen}. Then, we examine manually each of the caption that contains those words, and highlight the words that are used correctly in describing their respective image, either as correct object, action or attribute. The image-caption pair of some correctly inferred least seen words are shown in \figurename~\ref{fig:caption_eg} as examples. From Table \ref{least_seen}, we can see that our phrase-based model is generally able to infer correctly more words which are less seen, compared to the baseline in both Flickr30k and MS-COCO datasets. As for Flickr8k dataset, our baseline is able to infer the word `\textit{snowboarding}' which appears for 48 times in the training data, in the first image in \figurename~\ref{fig:caption_eg}. Nonetheless, our model has inferred `\textit{snowboarder}' for that image, which naturally makes the generation of action `\textit{snowboarding}' redundant. 
%, provided that there are significant differences between the caption generated with our model and baseline
%This is especially obvious in the MS-COCO dataset, where all correctly inferred least seen words has occurrence lower than that of baseline by a large margin. 

\begin{table*}
	\renewcommand{\arraystretch}{1.3}
	\caption{Top-5 most seen words that are not inferred in the generated captions.}
	\label{most_seen}
	\centering
	\begin{tabular}{c c| c c| c c| c c| c c| c c}
		\toprule
		\multicolumn{4}{c|}{\textbf{Flickr8k}}					  & \multicolumn{4}{c|}{\textbf{Flickr30k}}					  & \multicolumn{4}{c}{\textbf{MS-COCO}}						\\
		\hline
		\multicolumn{2}{c|}{Ours} & \multicolumn{2}{c|}{Baseline} & \multicolumn{2}{c|}{Ours} & \multicolumn{2}{c|}{Baseline} & \multicolumn{2}{c|}{Ours} & \multicolumn{2}{c}{Baseline}	\\
		\hline
		Words		& Seen		  & Words			& Seen		  & Words		& Seen		  & Words			& Seen		  & Words			& Seen	  & Words			& Seen			\\
		\hline
		while		& 1443		  & an				& 1807		  & up 			& 4762  	  & an				& 14590	 	  & by				& 16378	  & by				& 16378			\\
		child		& 1120		  & while			& 1443		  & as 			& 4598	  	  & one				& 5890		  & several			& 9082	  & there			& 12109			\\
		three		& 1052		  & child			& 1120		  & outside		& 4273		  & as				& 4598		  & sits			& 8847	  & three			& 10612			\\
		one			& 876		  & three			& 1052		  & from		& 3721		  & outside			& 4273		  & area			& 8377	  & several			& 9082			\\
		her			& 861		  & green			& 931 		  & their		& 3702	  	  & their			& 3702		  & one				& 8335	  & sits			& 8847			\\
		\bottomrule
	\end{tabular}
\end{table*}

\begin{figure*}[!t]
	\centering
	\includegraphics[height=0.25\linewidth, width=\linewidth]{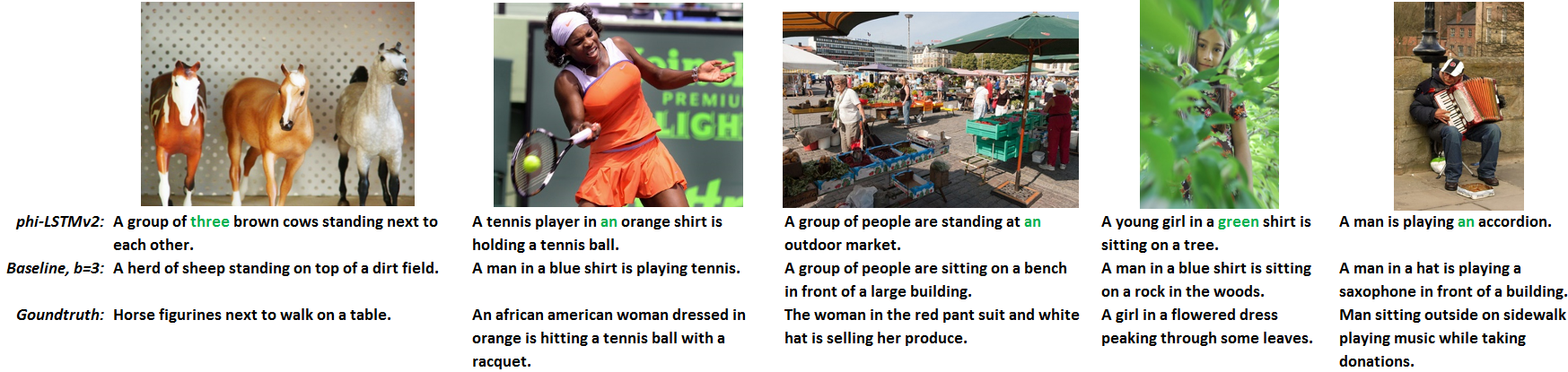}
	\caption{Examples of caption generated from image. The most seen words that are inferred by our model alone are in green.}\vspace{-.1in}
	\label{fig:caption_eg2}
\end{figure*}

Furthermore, we record the top five, {\it most seen} words which are absent in the generated captions of our model and the baseline in Table \ref{most_seen}. To have a better understanding of our findings, we group the list in Table \ref{most_seen} according to where the word usually appears in the caption. Starting with words that are salient in image, we notice that both models are not able to infer `\textit{sits}', '\textit{child}' and `\textit{several}', because the alternatives (e.g. `\textit{sitting}', `\textit{boy/girl}' and `\textit{group of}') are much more probable. The same goes for abstract scene such as `\textit{outside}' and `\textit{area}', where a definite scene description is more probable. As for attribute `\textit{green}', its inference is challenging in the Flickr8k dataset because there are a lot of green objects in the training data that are not described with the word `\textit{green}', such as grass, field, leaves etc. Although our model is able to infer `\textit{a green shirt}' in some cases, the image is actually people covered by leaves rather than people wearing a green shirt, as shown in the 4th image of \figurename~\ref{fig:caption_eg2}.

We also observe that our model is able to generate the word `\textit{an}' on all three datasets, while the baseline model can only do so in the MS-COCO dataset. One important reason is that the test set of MS-COCO dataset contains more objects starting with vowels (\textit{e.g. elephant}), while there are very few of such cases in both of the Flickr datasets. Nonetheless, there are still attributes starting with vowels such as `\textit{an orange shirt}' and `\textit{an outdoor market}' in both Flickr datasets, as shown in \figurename~\ref{fig:caption_eg2}. We note that generating caption in a phrase-based manner increases the chance of their inference and retain during beam search. Caption generated in a pure sequential model would result in word `\textit{a}' being inferred first due to the local statistic of data, and this will greatly reduce the chance of attribute or object starts with vowel being inferred next. The same applies for the word `\textit{there}' in the MS-COCO dataset. Since our AS decoder does not have GTS with word `\textit{a}' as the first word, we have a better chance of generating caption starting with `\textit{there}'. This is one the reasons that our model is capable of generating more unique captions compared to the baseline. 

Possessive pronoun such as `\textit{her}' and `\textit{their}' are not inferred by both models\footnote{Word `\textit{her}' is the 7\textit{th} most seen word absent from the generated captions of baseline model in Flickr8k dataset.} in the Flickr datasets because they usually appear before the human body parts (\textit{e.g.hand, head}) which are not as salient as the human, or appear before pets (\textit{e.g. dog}) where words `\textit{a}' gains a higher probability during inference. Nevertheless, a dataset as large as the MS-COCO would solve this problem. On the other hand, cardinal number `\textit{three}' has much less chances to obtain a high probability score than `\textit{a}' or `\textit{two}'. Still, our model can infer word `\textit{three}' in the MS-COCO dataset, as shown in first image of \figurename~\ref{fig:caption_eg2}. 

We also found that word `\textit{by}' is used mostly as preposition in the MS-COCO dataset, but `\textit{next to}', `\textit{in front of}' and `\textit{near}' are the preferable alternatives for both models. Other non-visual words such as particle `\textit{up}' and conjunction `\textit{from}' have better chance to be inferred by the baseline model, as a result of longer generated caption. However, both models are still incapable of inferring conjunction `\textit{while}' and `\textit{as}', which are mostly used to describe multiple actions performed by the same or different individuals. The same applies to word `\textit{one}'\footnote{Word `\textit{one}' is the 6\textit{th} most seen word absent from the generated captions of baseline model in the Flickr8k and MS-COCO datasets.}, which is mostly used to describe different individuals within a group (\textit{e.g. Two dogs, one black, one brown...}). We reason that sentence of such structure might require attention mechanism on the image to generate, though no researchers in this field has reported that a model of such capability is developed to the best of our knowledge.

\section{Conclusion}
We have presented a phrase-based LSTM (phi-LSTM) model to generate image caption in a hierarchical manner, where NPs that describe the salient objects in an image are first generated, before a complete caption is formed from the NPs. Each generated NP is encoded as a compositional vector, which acts as the input of one time step at the sentence level. Such design allows NPs to be decoded in a consistent time-scale, while reducing the variation of time-scale at the sentence level. Empirical results show that image caption generated in such manner is more precise in terms of object and attribute, when compared with a pure sequential model using words as atomic unit. Moreover, the hierarchical decoding process allows more novel captions with diverse word content to be generated. Our future work will focus on designing of a phrase-based bi-directional model for image captioning.  

\bibliographystyle{IEEEtran}
\bibliography{egbib}

\newpage

\begin{figure*}[!h]
	\centering
	\includegraphics[height=0.22\linewidth, width=\linewidth]{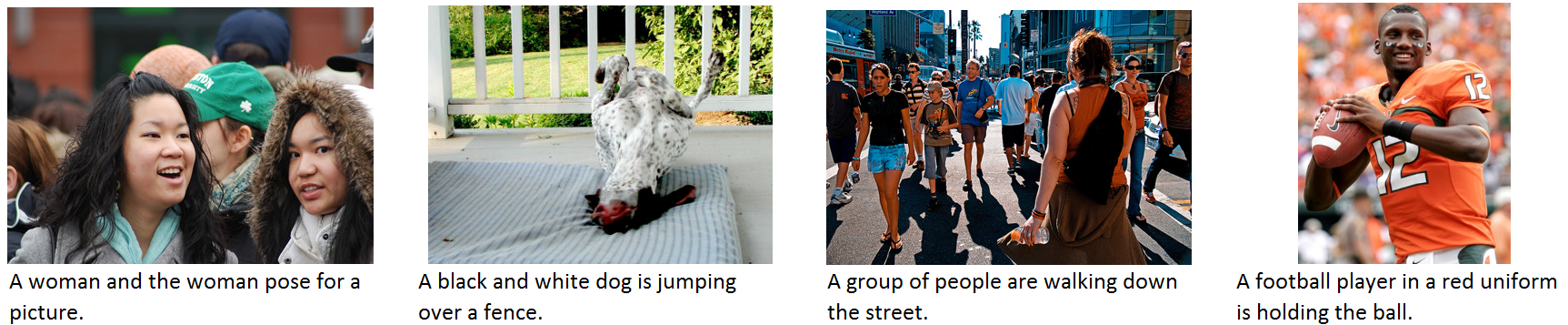}
\end{figure*}
\begin{figure*}[!h]
	\centering
	\includegraphics[height=0.22\linewidth, width=\linewidth]{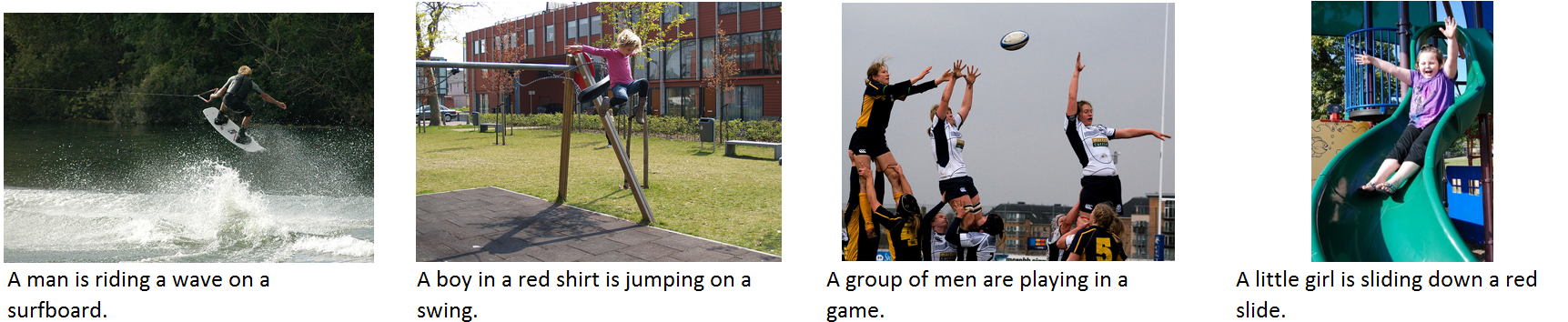}
\end{figure*}
\begin{figure*}[!h]
	\centering
	\includegraphics[height=0.22\linewidth, width=\linewidth]{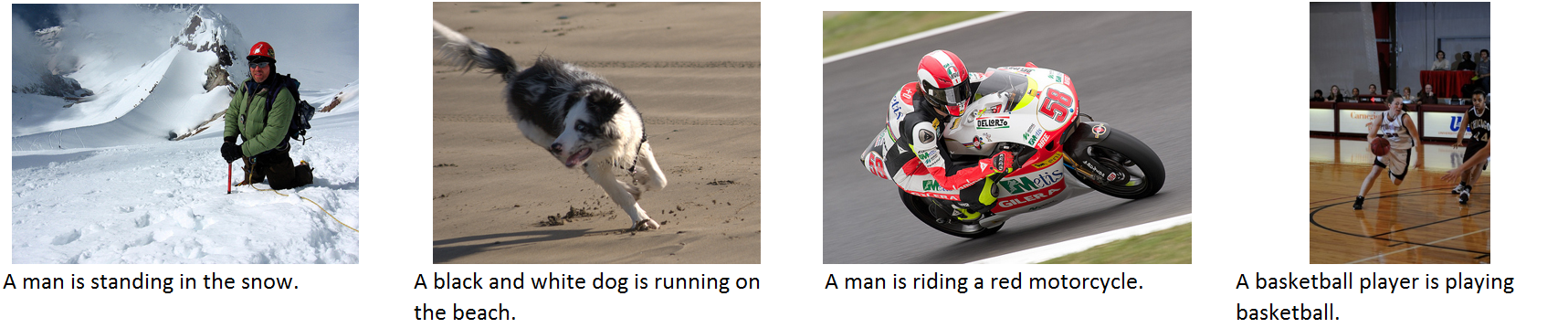}
\end{figure*}
\begin{figure*}[!h]
	\centering
	\includegraphics[height=0.22\linewidth, width=\linewidth]{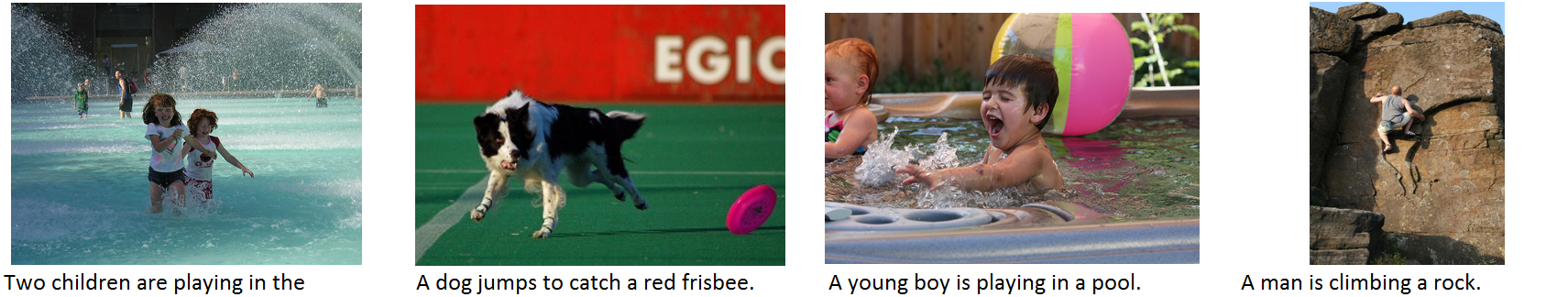}
\end{figure*}
\begin{figure*}[!h]
	\centering
	\includegraphics[height=0.22\linewidth, width=\linewidth]{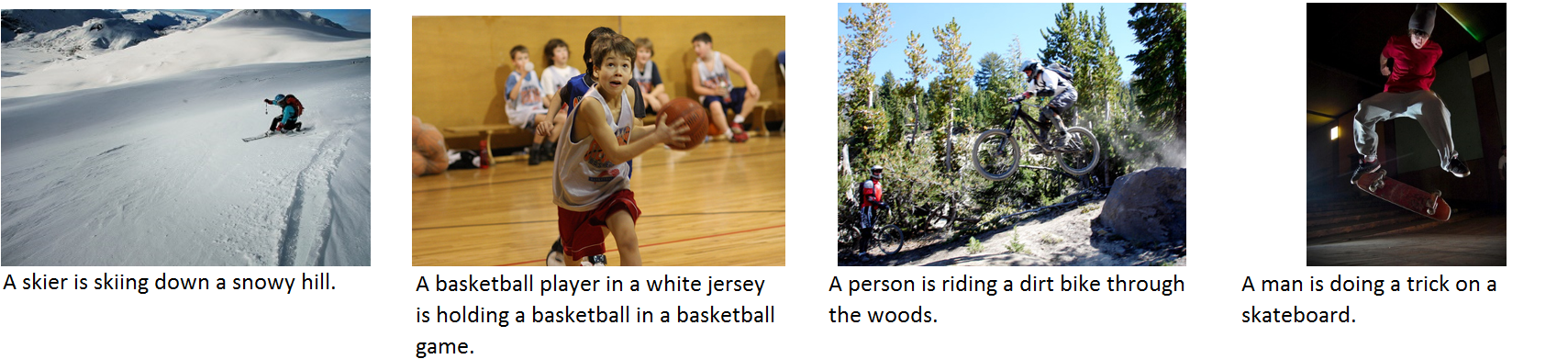}
	\caption{Flickr8k dataset: Sample image captioning results.}
\end{figure*}

\begin{figure*}[!h]
	\centering
	\includegraphics[height=0.22\linewidth, width=\linewidth]{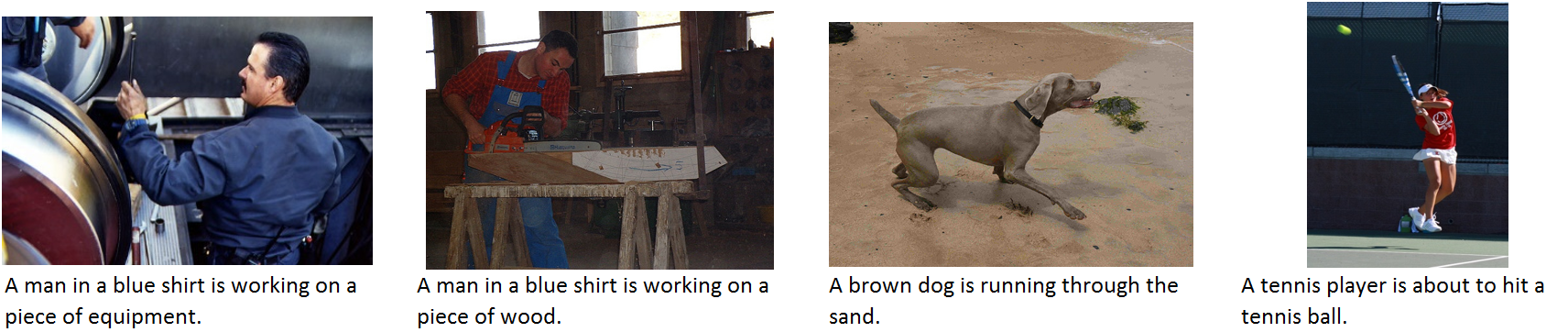}
\end{figure*}
\begin{figure*}[!h]
	\centering
	\includegraphics[height=0.22\linewidth, width=\linewidth]{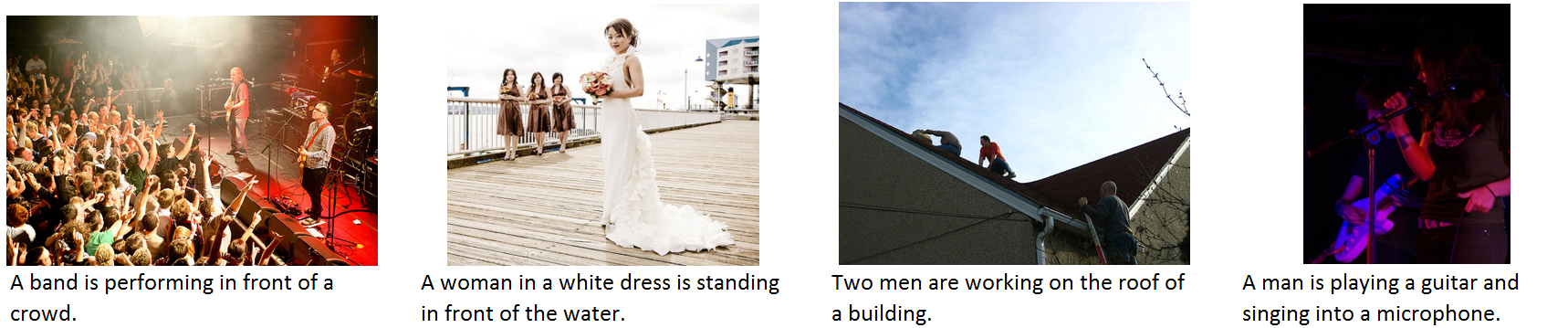}
\end{figure*}
\begin{figure*}[!h]
	\centering
	\includegraphics[height=0.22\linewidth, width=\linewidth]{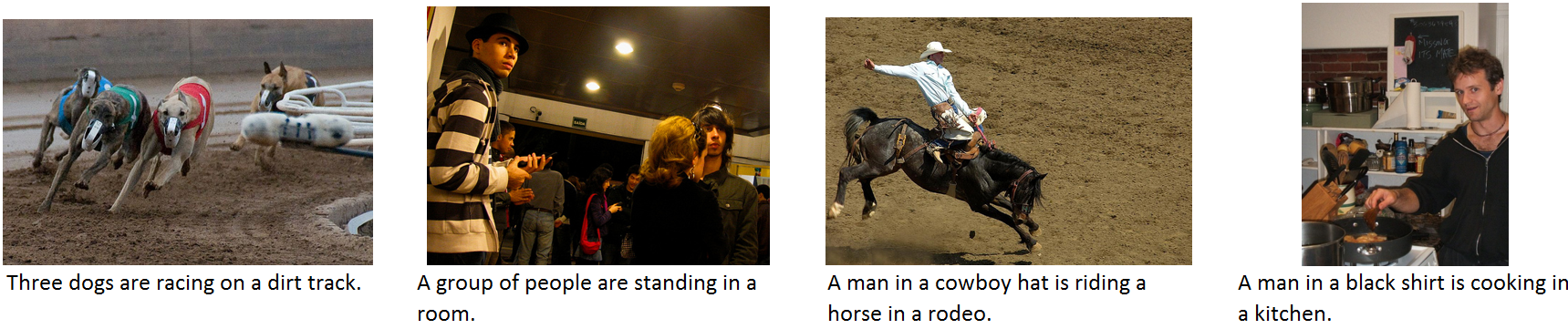}
\end{figure*}
\begin{figure*}[!h]
	\centering
	\includegraphics[height=0.22\linewidth, width=\linewidth]{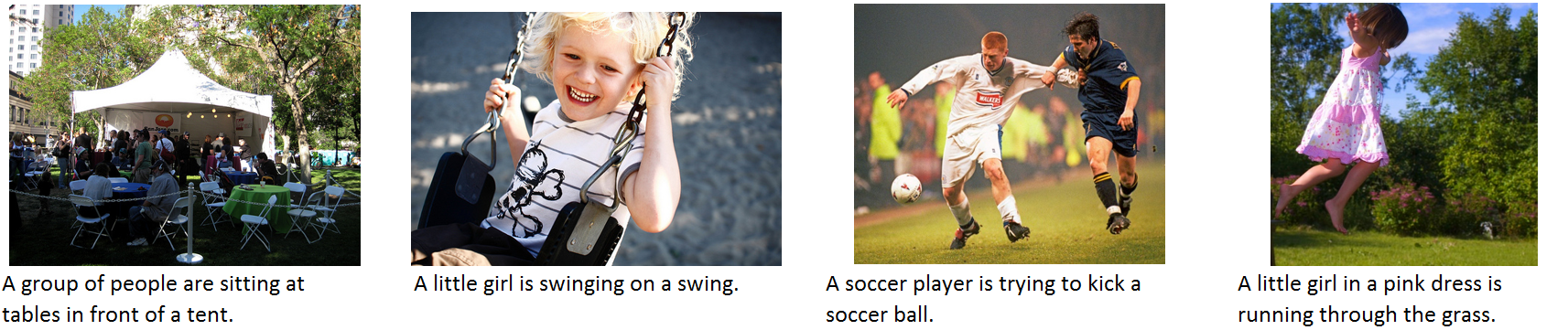}
\end{figure*}
\begin{figure*}[!h]
	\centering
	\includegraphics[height=0.22\linewidth, width=\linewidth]{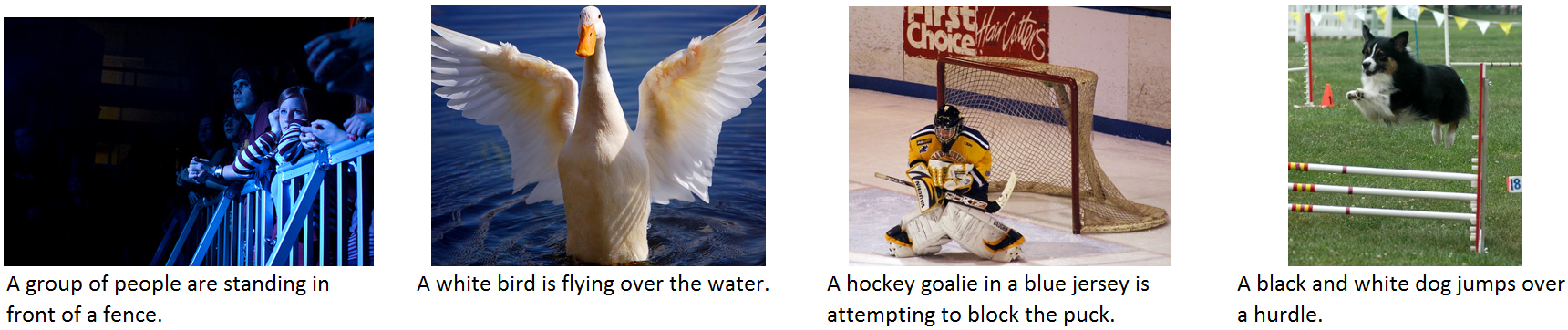}
	\caption{Flickr30k dataset: Sample image captioning results.}
\end{figure*}

\begin{figure*}[!h]
	\centering
	\includegraphics[height=0.22\linewidth, width=\linewidth]{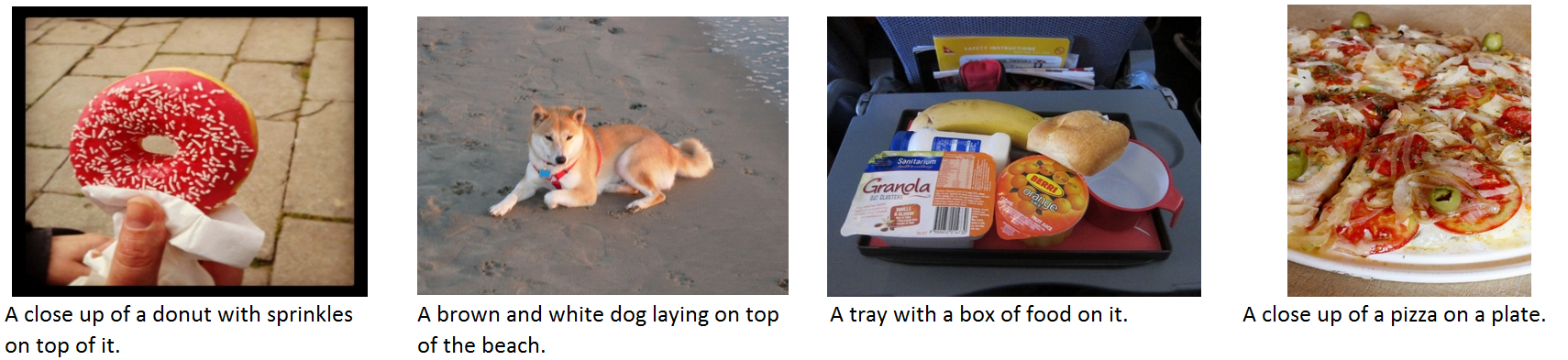}
\end{figure*}
\begin{figure*}[!h]
	\centering
	\includegraphics[height=0.22\linewidth, width=\linewidth]{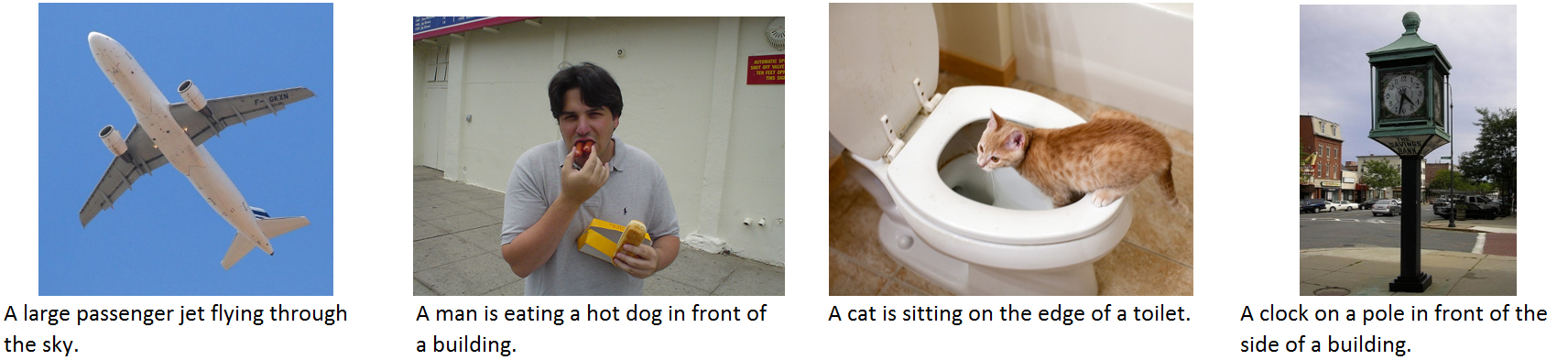}
\end{figure*}
\begin{figure*}[!h]
	\centering
	\includegraphics[height=0.22\linewidth, width=\linewidth]{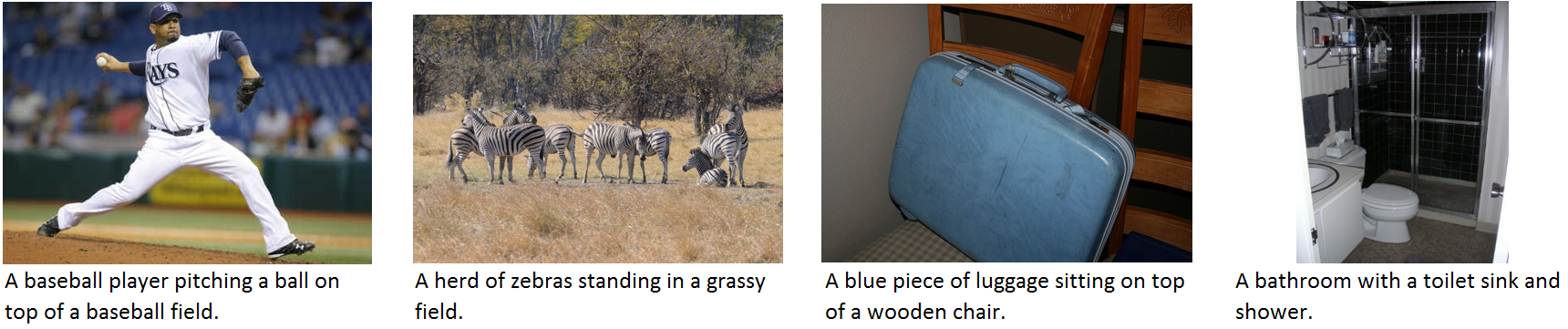}
\end{figure*}
\begin{figure*}[!h]
	\centering
	\includegraphics[height=0.22\linewidth, width=\linewidth]{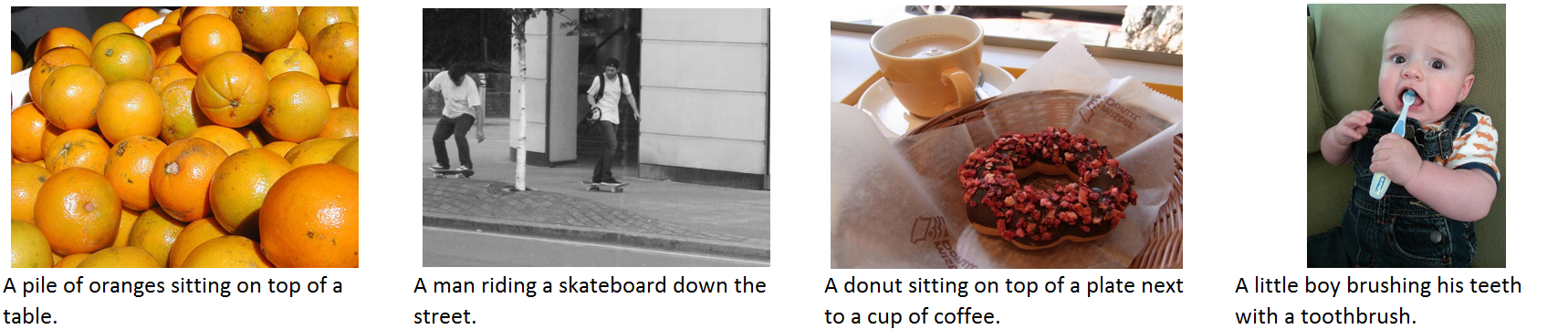}
\end{figure*}
\begin{figure*}[!h]
	\centering
	\includegraphics[height=0.22\linewidth, width=\linewidth]{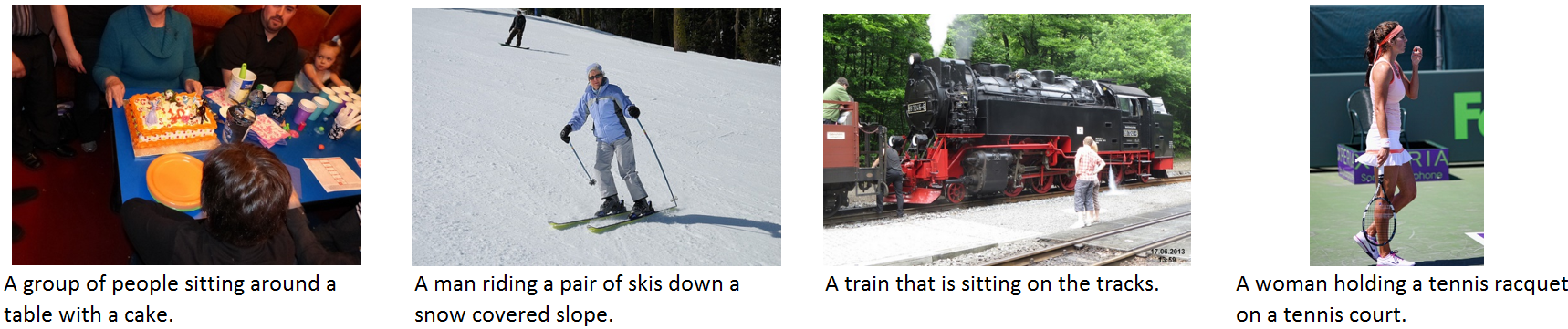}
	\caption{MS-COCO dataset: Sample image captioning results.}
\end{figure*}

\vfill


\end{document}